\documentclass[runningheads]{llncs}
\usepackage[T1]{fontenc}
\usepackage{graphicx}
\usepackage{booktabs}
\usepackage[misc]{ifsym}
\newcommand{\corr}{(\Letter)}

\usepackage{geometry}
\usepackage{xcolor}
\definecolor{winter}{rgb}{0.85,0.08,0.2}
\definecolor{summer}{rgb}{0.95,0.53,0.18}         
\definecolor{spring}{rgb}{0.02,0.93,0.68}
\definecolor{autumn}{rgb}{0.02,0.68,0.9}
\definecolor{rightblue}{RGB}{76,114,176} 
\definecolor{green(munsell)}{rgb}{0.0, 0.66, 0.47} 

\usepackage{amsmath}
\usepackage{amssymb}
\usepackage{hyperref}
\hypersetup{
    colorlinks=true,
    citecolor=autumn,
    linkcolor=winter,
    filecolor=magenta,      
    urlcolor=cyan,
    }

\usepackage{booktabs}
\usepackage{siunitx}
\usepackage{multirow}
\usepackage{graphicx}
\usepackage{ulem}

\usepackage{pgfplots}
\pgfplotsset{compat=1.18}

\begin{document}


\title{SpecMoE: Spectral Mixture-of-Experts Foundation Model for Cross-Species EEG Decoding}

\titlerunning{EEG Foundation Model - SpecMoE}

\author{Davy Darankoum\inst{1,2} \and
Chlo\'e Habermacher\inst{2} \and
Julien  Volle\inst{2}* \corr \and
Sergei Grudinin\inst{1}* \corr}


\authorrunning{D. Darankoum et al.}

\institute{Univ. Grenoble Alpes, CNRS, Grenoble INP, LJK 38000 Grenoble, France 
\and
SynapCell SAS, ZAC ISIPARC, 38330 Saint-Ismier, France \\
* These authors jointly supervised the work \\
\email{jvolle@synapcell.fr} \\
\email{sergei.grudinin@univ-grenoble-alpes.fr}}

\maketitle              

\begin{abstract}
Decoding the orchestration of neural activity in electroencephalography (EEG) signals is a central challenge in bridging neuroscience with artificial intelligence. Foundation models have made strides in generalized EEG decoding, yet many existing frameworks primarily relying on separate temporal and spectral masking of raw signals during self-supervised pretraining. Such strategies often tend to bias learning toward high-frequency oscillations, as low-frequency rhythmic patterns can be easily inferred from the unmasked signal.

We introduce a foundation model that utilizes a novel Gaussian-smoothed masking scheme applied to short-time Fourier transform (STFT) maps. By jointly applying time, frequency, and time-frequency Gaussian masks, we make the reconstruction task much more challenging, forcing the model to learn intricate neural patterns across both high- and low-frequency domains.  To effectively recover signals under this aggressive masking strategy, we design SpecHi-Net, a U-shaped hierarchical architecture with multiple encoding and decoding stages. To accelerate large-scale pretraining, we partition the data into three subsets, each used to train an independent expert model. We then combine these models through SpecMoE, a mixture of experts framework guided by a learned spectral gating mechanism.

SpecMoE achieves state-of-the-art performance across a diverse set of EEG decoding tasks, including sleep staging, emotion recognition, motor imagery classification, abnormal signal detection, and drug effect prediction. Importantly, the model demonstrates strong cross-species and cross-subject generalization, maintaining high accuracy on both human and murine EEG datasets. These findings suggest that spectral-aware Gaussian-smoothed masking combined with hierarchical feature integration provides a powerful inductive bias for next-generation EEG foundation models and brain–computer interface systems.


\end{abstract}

\section{Introduction}
Electroencephalography (EEG) remains the primary non-invasive gateway to understanding human cortical dynamics, offering superior temporal resolution for clinical diagnostics and Brain-Computer Interface (BCI) development. However, the inherent challenges of EEG: high non-stationarity, low signal-to-noise ratio, and significant inter-subject variability, have historically confined decoding models to narrow, task-specific applications.

The emergence of EEG foundation models has recently shifted the decoding paradigm toward large-scale self-supervised pretraining \cite{kuruppu2025eeg,shen2026brain4fms,liu2026eeg}. By leveraging masked signal reconstruction, these models learn generalized representations across massive unlabeled datasets. However, current frameworks predominantly apply rectangular masks to raw temporal or spectral data independently. This approach suffers from two critical flaws.
First, sharp masking boundaries introduce high-frequency edge artifacts that bias the model toward "discontinuity recovery" in addition to endogenous neural features. In the context of EEG, this is particularly problematic, as neural oscillations are fundamentally smooth and rhythmic. Forcing the model to reconstruct artificial transients at every mask boundary acts as a spectral interference, diverting a portion of the optimization process away from the underlying neural manifold. 
Second, standard masking strategies are susceptible to information leakage in the low-frequency domain. Because these slow oscillations span across long temporal windows, the model can straightforwardly infer the low-frequency patterns from the unmasked EEG segments. Consequently, the pretraining task becomes trivial in the lower bands, preventing foundation models from grasping the critical long-range rhythmic structures that characterize many cognitive and pathological states.

\begin{figure}[t]
    \centering
    \includegraphics[width=1\linewidth]{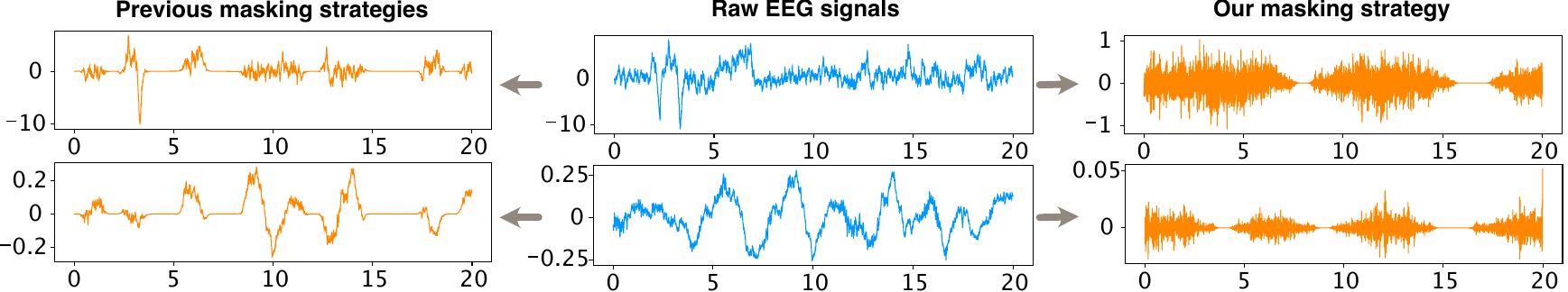}
    \caption{Left: previous masking strategies, where rectangular masks remove some time frames. Middle: original EEG signals. Right: proposed Gaussian masks remove some frequency oscillations (mostly low frequencies) in addition to certain time frames.
    The horizontal axis represents time in seconds, and the vertical axis is voltage in $\mu$V.}
    \label{fig:fig1}
\end{figure}

To address these limitations, we introduce SpecMoE, a spectral-anchored foundation model. Unlike previous methods, SpecMoE operates in the time-frequency domain using a novel Gaussian-smoothed masking strategy on short-time Fourier transform  (STFT) spectrograms (see Fig. \ref{fig:fig1} for a comparison with other masking strategies). By replacing sharp boundaries with smooth transitions, we eliminate the bias toward non-physiological transients, ensuring that the optimization process remains focused on physiological neural rhythms. Furthermore, our joint time and frequency masking geometry, where specific frequency bands are masked over the entire duration of the EEG segment, prevents any low-frequency leakage between the masked and remaining signals. Such masking geometry forces the model to learn the complex long-range dependencies of the neural manifold rather than relying on simple interpolation. 
To effectively recover signals under this more challenging masking regime, we design SpecHi-Net, a U-shaped hierarchical architecture. By utilizing multiple encoding and decoding stages, SpecHi-Net captures multi-scale temporal and spectral features, providing the structural depth necessary for high-fidelity reconstruction where shallower, non-hierarchical transformers typically fail.
Finally, to improve domain-specific adaptation on downstream applications,
we propose a Spectral-guided Mixture of Experts (SpecMoE) framework. Governed by a learned spectral gating mechanism that routes information based on the signal’s Power Spectral Density (PSD), our model dynamically weights expert contributions to match the rhythmic content of the task at hand. This approach enables state-of-the-art performance and robust cross-species generalization across nine heterogeneous benchmarks.
In summary, our work comprises the following main contributions:
\begin{itemize}
    \item A novel Gaussian-smoothed masking strategy for self-supervised pretraining
    \item An architecture with emphasis on multi-level encoder-decoder (SpecHi-Net)
    \item A novel spectral gating mechanism for a Mixture of Experts framework
    \item A novel cross-species validation approach for EEG foundation models
\end{itemize}

The code and pretrained models are available at \url{https://github.com/TeraXj78/SpecMoE}.

\section{Related Work}

\subsubsection{Recent Evaluations}
EEG-FM-Bench is a recent comprehensive resource that compares foundation EEG models on 14 open-source datasets across 10 common EEG
paradigms \cite{xiong2025eeg}. According to the benchmark, EEGPT \cite{wang2024eegpt} and CBraMod \cite{wangcbramod} are the top-performing models, being both compact and effective. The authors also emphasize the importance of fine-tuning with all parameters from the pretrained weights, and critically evaluate single-task versus multi-task fine-tuning objectives.
Liu et al. \cite{liu2026eeg}  recently provided another evaluation of 12 open-source foundation models and competitive task-specific baselines across 13 EEG datasets spanning nine BCI paradigms. 
They also concluded that full-parameter fine-tuning is beneficial and noted that task-specific models can still be competitive. Overall, CBraMod \cite{wangcbramod} was again the top-performing foundation model in this study, while
EEGNet \cite{lawhern2018eegnet} demonstrated the best performance among task-specific models. Brain4FMs \cite{shen2026brain4fms}  and Kuruppu et al. \cite{kuruppu2025eeg} are yet another recent attempts to benchmark foundation models, but they did not include some state-of-the-art approaches, e.g., CSBrain. Interestingly, the event-related potential (ERP) benchmark revealed that the EEG foundation models do not outperform the supervised task-specific baselines in ERP tasks \cite{wang2026benchmarking}, where EEGConformer \cite{kasthuri2024eeg} achieved the highest accuracy for ERP classification.

\subsubsection{Task-Specific and Foundation Models}
Despite the emergence of foundation EEG models, smaller, task-specific architectures continue to serve as strong baselines. 
For instance, EEGNet \cite{lawhern2018eegnet} is a lightweight CNN-based model that utilizes depthwise and separable convolutions to extract robust features. FFCL \cite{li2022motor} combines parallel CNN and LSTM branches, allowing CNNs to extract spatial features while LSTMs capture temporal dynamics. The features from these multiple branches are then fused using a fully connected layer. Another notable task-specific baseline is EEGConformer \cite{kasthuri2024eeg}, which employs self-attention mechanisms to capture long-range temporal dependencies and global contextual information.

LaBraM \cite{jianglarge} is one of the pioneering EEG foundation models, pre-trained on 2,500 hours of data using vector quantized variational autoencoders for dual frequency-phase mask learning. CBraMod \cite{wangcbramod} introduces criss-cross attention to separately capture spatial and temporal features within the same transformer layer, enhancing its representational capabilities. CSBrain \cite{zhou2025csbrain} builds upon CBraMod and incorporates cross-scale spatiotemporal tokenization along with structured sparse attention to create robust EEG representations. A novel decoder-centric paradigm, inspired by developments in large language models (LLMs), was introduced in ECHO \cite{liu2025echo}. LEAD \cite{wang2025lead} uses separate temporal and spatial attentions followed by a learnable gating  fusion. During pretraining, it employs one of five augmentation strategies,  including individual temporal, frequency, or channel masking. REVE \cite{elreve} introduced 4D spatial-temporal positional encoding with massive-scale pretraining. Similarly, DeeperBrain  \cite{wang2026deeperbrain} leverages the spatial and temporal organization of the data by modeling 3D electrode geometry and a learnable spatial decay kernel. It also introduces neurodynamics statistics learning during pretraining. UNI-NTFM \cite{chen2025uni} is currently the largest EEG foundation model, featuring up to 1.9 billion parameters. It encodes time, frequency, and raw signal representations separately, followed by cross-attention. Additionally, it introduces spatial priors into input features and optimizes fine-tuning using a mixture-of-experts neural transformer.

The most popular pretraining objective comprises randomly masked time-domain signal reconstruction \cite{liu2026eeg}. Yet, classical BrainBERT \cite{wangbrainbert} and BrainWave \cite{yuan2024brainwave} approaches are pretrained on masked STFT spectrograms, while BioCodec \cite{avramidis2025neural} employs an STFT-based spectral loss computed at multiple scales, EEGFormer \cite{wan2023eegformer} reconstructs spectral amplitudes, and ALFEE \cite{xiong2025alfee} reconstructs the PSD. Notably, TFM-Tokenizer \cite{pradeepkumar2025tokenizing} introduces an explicit dual-path time-frequency rectangular masking objective to disentangle temporal and spectral motifs, encouraging the model to learn frequency-specific patterns across time.

The Mixture of Experts (MoE) framework is gaining traction in EEG foundation models. 
The MGEC model \cite{zhang2026mutual} combines shared and routed experts, and incorporates a MoE architecture for the latter. 
UNI-NTFM \cite{chen2025uni} also applies the MoE concept during fine-tuning optimization. 
Additionally, BrainMoE \cite{mabrainmoe}  seeks to develop a channel-wise MoE structure.

\section{SpecMoE Framework}

\subsubsection{Pretraining Design}
\begin{figure}[t]
    \centering
    \includegraphics[width=1\linewidth]{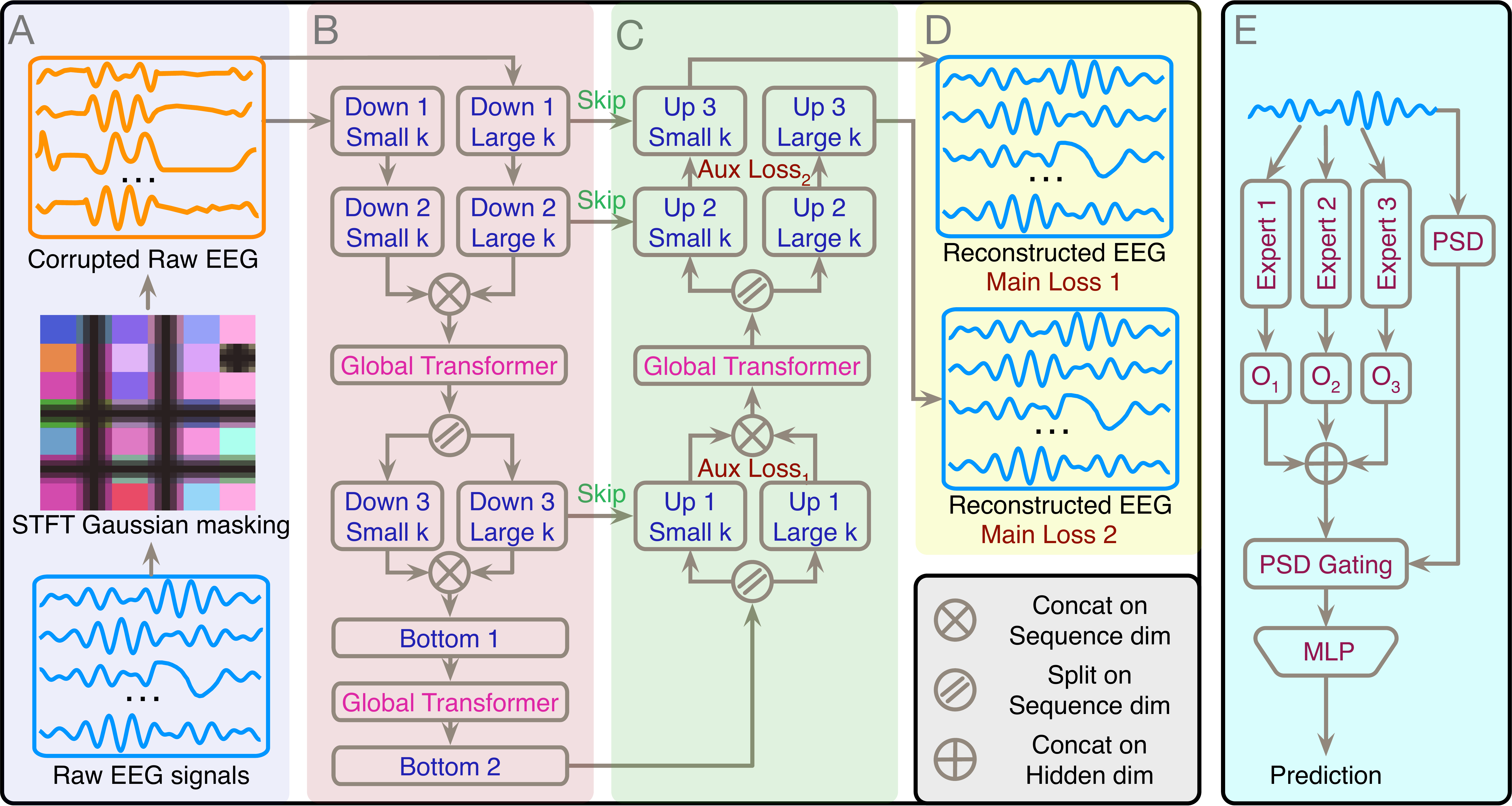}
    \caption{SpecMoE overview. A) Gaussian-based masking pipeline. B) Hierarchical encoder, with 'k' standing for 'kernel'. C) Hierarchical decoder. D) Reconstruction objective. E) Fine-tuning pipeline.}
    \label{fig:fig2}
\end{figure}

SpecMoE pretraining (Fig. \ref{fig:fig2}) follows a self-supervised generative paradigm designed to reconstruct signals from spectral-temporal corruption.
The pretraining pipeline is structured into three phases: EEG signal corruption, latent space learning, and multi-scale signal reconstruction.

\paragraph{Phase I: EEG Signal Corruption.} 
The input $\textbf{x}\in \mathbb{R}^{C \times L}$, where $C$ denotes the number of recording channels and $L$ denotes the sequence length (total number of time steps), is first projected into the time-frequency domain via an STFT, yielding the complex-valued representation $\textbf{z} \in \mathbb{C}^{C \times F \times T}$. 
Here, $F$ and $T$ represent the frequency bins and temporal frames, respectively. 
We then apply a proposed {\em Gaussian-smoothed masking strategy}  
to $\textbf{z}$, designed to prevent spectral leakage and preserve low-frequency physiological rhythms. 
The corrupted temporal signal $\hat{\textbf{x}} \in \mathbb{R}^{C \times L}$ is subsequently recovered through an inverse STFT (iSTFT), serving as the input for the neural backbone.

\paragraph{Phase II: Hierarchical Latent Space Encoding.} 
The corrupted signal $\hat{\textbf{x}}$ is processed by {\em SpecHi-Net}, a {\em Spectral-Hierarchical Network} that adopts a U-Net-like hierarchical structure to enable multi-scale feature extraction and high-fidelity signal recovery. The encoder of SpecHi-Net consists of three successive downsampling stages ($Down_1$, $Down_2$, and $Down_3$), each employing a {\em dual-path convolutional encoder}
to capture simultaneously transient micro-states and long-range oscillatory patterns. To capture global dependencies across channels,
these convolutional features are interleaved with  {\em Global Transformer}
layers utilizing Rotary Positional Encodings  (RoPE) \cite{su2024roformer}. 
We specifically do not encode channel absolute positions, processing them equally by the architecture, which ensures its channel-size invariance.

\paragraph{Phase III: Multi-Scale EEG signal Reconstruction.} The symmetric decoder of SpecHi-Net integrates the latent representations learned by the encoder, through a series of upsampling stages ($Up_1$, $Up_2$, and $Up_3$) and skip-connections. To enforce structural and physiological consistency, the decoder generates reconstructions at multiple levels of granularity: one intermediate signal reconstructed after $Up_1$ layer ($\text{\textbf{interm-}}\textbf{x}_\textbf{1}$), a second intermediate signal reconstructed after $Up_2$ layer ($\text{\textbf{interm-}}\textbf{x}_\textbf{2}$) and finally two reconstructions from $Up_3$ outputs ($\tilde{\textbf{x}}_\textbf{1}$, $\tilde{\textbf{x}}_\textbf{2}$). The entire system is optimized via a  {\em multi-objective loss function}
that minimizes error in both the temporal and spectral domains.

\subsubsection{Gaussian-Smoothed Masking Strategy}
To compel SpecHi-Net to learn the intrinsic spectral-temporal dynamics of neural signals, we propose a novel {\em Gaussian-smoothed masking strategy} applied in the time-frequency domain. Unlike traditional temporal masking, which creates sharp discontinuities and high-frequency edge artifacts, our approach utilizes STFT coefficients to apply "soft" masks that respect the biological rhythmic structures of the EEG (see Fig. \ref{fig:fig1}).

Given an input EEG signal $\textbf{x} \in \mathbb{R}^{C \times L}$, we first compute its STFT representation $\textbf{z} \in \mathbb{C}^{C \times F \times T}$, where $F$ and $T$ denote the frequency and time-frame dimensions, respectively. 
We define a 2D mask map $\textbf{M}(f, t) \in [0, 1]^{F \times T}$, initialized as a matrix of ones. We iteratively subtract Gaussian kernels from $\textbf{x}$ until a target mask ratio $\rho = 0.5$ is reached. A single Gaussian kernel $G(f, t)$ centered at coordinates $(f_0, t_0)$ is defined as:

\begin{equation}
G(f, t) = \exp\left( -\frac{(f - f_0)^2}{2\sigma_f^2}  \right)
\exp\left(  - \frac{(t - t_0)^2}{2\sigma_t^2} \right),
\end{equation}
where  we set parameters $\sigma_f$ and $\sigma_t$ 
to $5\%$ of the total range of the frequency domain
$F$ and time domain  $T$, respectively.
The mask map is updated via:
\begin{equation}
\mathbf{M}(f, t) \leftarrow \mathbf{M}(f, t) \cdot (1 - G(f, t)).
\end{equation}
The same mask $\textbf{M}$ is applied to all the channels C of the STFT representation $\textbf{z}$. Finally, the masked signal $\hat{\textbf{x}}$ is generated via the inverse STFT: 
\begin{equation}
\hat{\textbf{x}} = \text{iSTFT}(\textbf{z} \odot \mathbf{M}).
\end{equation}

\subsubsection{Multi-Type Masking Strategy} 
We simultaneously employ three distinct masking geometries governed by a probability distribution $P = [0.6, 0.3, 0.1]$ for frequency, time, and joint time-frequency masking, respectively:
    
    \paragraph{Frequency Masking ($\sigma_t \to \infty$):} 
    This mode obscures specific spectral bands across the entire temporal window, compelling SpecHi-Net to reconstruct frequency-domain information from broader cross-temporal correlations. By extending the mask across the full duration of the segment, we effectively eliminate \textit{spectral information leakage}; since a masked frequency bin is unavailable at any time point, the network is forced to learn the underlying physiological dependencies between different neural oscillations.
    
    \paragraph{Time Masking ($\sigma_f \to \infty$):} 
    This mode removes contiguous temporal segments across the entire spectrum. By obscuring all frequency components for a given time frame, the strategy forces the model to capture the long-range temporal dynamics of the signal. SpecHi-Net must therefore learn to interpolate transient neural events by leveraging the temporal continuity and phase consistency of the preceding and succeeding spectral contexts.
    
    \paragraph{Joint Time-Frequency Masking:} 
    This mode utilizes localized 2D Gaussian "blobs" to obscure specific regions in the time-frequency plane. Unlike the previous modes, this geometry targets discrete neuro-spectral events. By masking localized clusters of energy, it challenges the model to perform joint spectral-temporal inference, reconstructing the missing information by simultaneously utilizing neighboring frequency bands and adjacent time steps. This simulates the recovery of obscured biological biomarkers while promoting a robust, multi-dimensional latent representation.

\subsubsection{Spectral-Band Bias} 
A critical innovation of SpecMoE is the {\em band-informed bias}. 
During the selection of frequency centers $f_0$, we specifically constrain $50\%$ of the masks to correspond to the primary EEG physiological bands: $\delta$ (1--4 Hz), $\theta$ (4--8 Hz), $\alpha$ (8--12 Hz), and $\beta$ (12--30 Hz). 
By prioritizing the masking of these low-frequency rhythmic structures, 
which are often not targeted in previous masking strategies, we ensure that our model develops a high-fidelity understanding of a broad range of clinically relevant biomarkers.

\subsubsection{Hierarchical Dual-Path Convolutional Encoder}
To address the multi-scale nature of EEG signals, SpecHi-Net utilizes a hierarchical encoder composed of three primary levels ($Down_1$, $Down_2$, and $Down_3$). Following previous works \cite{darankoum2025cosupformer,eldele2021attention}, each stage features a \textit{dual-path} architecture: one path employs small-kernel convolutions for transient detection, while the parallel path utilizes large-kernel, dilated convolutions to capture long-range rhythmic structures. This design ensures that both localized micro-states and global oscillatory patterns are encoded into the latent space.

\subsubsection{Interleaved Global Transformer}
We inserted three transformer layers interleaved at hierarchical transitions and within the architectural bottleneck ($Bottom_1$ and $Bottom_2$) utilizing Multi-Head Attention (MHA). To ensure the learned representations remain robust to the varying sequence lengths encountered across diverse downstream tasks, we adopt Rotary Positional Encodings (RoPE) \cite{su2024roformer}. We strategically selected RoPE over more traditional absolute positional embeddings. Indeed, it allows the model to maintain temporal consistency across the 1s to 60s windows found in our downstream datasets and ensures resolution invariance as the signal is compressed through the hierarchy of SpecHi-Net.

A distinguishing feature of our transformer implementation is the processing of the dual-path encoder outputs. Before entering the attention blocks, the feature maps from the short-range and long-range convolutional paths are flattened and concatenated along the temporal dimension. This operation enables the transformer to: {\em (1) Learn cross-feature relationships:} By observing features from both small ($k=4$) and wide ($k=65$) kernels simultaneously, the model can learn how localized transients (e.g., spikes) correlate with broader rhythmic oscillations (e.g., slow waves). {\em (2) Capture global dependencies:} Operating on the flattened channel-time representation allows the model to integrate context across the entire electrode manifold.

Since the Gaussian mask is applied uniformly across all channels, this architecture compels the model to reconstruct obscured spectral-temporal regions by modeling the complex oscillations inherent in large-scale brain networks rather than relying on local temporal redundancies.

\subsubsection{Multi-Objective Loss Design}
To ensure that SpecHi-Net captures both the high-fidelity temporal morphology and the essential spectral characteristics of EEG signals, we optimize the framework using a composite loss function $\mathcal{L}_{total}$. This function integrates multi-scale temporal reconstruction errors with a channel-wise spectral penalty.
%
For any predicted signal $\textbf{p}$ and its corresponding ground truth $\textbf{y}$, we define a base composite loss $\mathcal{L}(\textbf{p}, \textbf{y})$ that balances time-domain mean squared error (MSE) with frequency-domain consistency:
\begin{equation}
\mathcal{L}(\textbf{p}, \textbf{y}) = \text{MSE}(\textbf{p}, \textbf{y}) + w_{spec}  \mathcal{L}_{spec}(\textbf{p}, \textbf{y}),
\end{equation}
where $w_{spec} = 0.02$ to reach the same order of magnitude as the temporal loss. We designed the spectral loss component $\mathcal{L}_{spec}$ with a specific goal to be sensitive to the reconstructed signal amplitudes $|\text{STFT}(\textbf{p}_c)|$ but not the phases,
\begin{equation}
\mathcal{L}_{spec}(\textbf{p},\textbf{y}) = \frac{1}{C} \sum_{c=1}^{C} \left( |\text{STFT}(\textbf{p}_c)| - |\text{STFT}(\textbf{y}_c)| \right)^2 .
\end{equation}
This spectral regularizer ensures that the model preserves the biological power distribution across frequency bands, even when temporal alignments are subtly shifted.
%
To facilitate stable training and enforce structural consistency across the hierarchical depth of the decoder, we utilize multi-level supervision. The total loss $\mathcal{L}_{total}$ combines the primary dual-path reconstruction losses with intermediate auxiliary losses:
\begin{equation}
\mathcal{L}_{total} = \sum_{i \in {1,2}} \mathcal{L}(\tilde{\textbf{x}}_\textbf{i}, \textbf{x}) + \sum_{j \in {1,2}} \alpha_j \mathcal{L}(\text{\textbf{interm-}}\textbf{x}_\textbf{j}, \textbf{x}{\downarrow}),
\end{equation}
where $\tilde{\textbf{x}}_\textbf{1}$ and $\tilde{\textbf{x}}_\textbf{2}$ are the final outputs from $Up_3$, $\text{\textbf{interm-}}\textbf{x}_\textbf{1}$ is the intermediate output from $Up_1$ and $\text{\textbf{interm-}}\textbf{x}_\textbf{2}$ is the intermediate output from $Up_2$. For auxiliary supervision, the ground truth $\textbf{x}$ is dynamically downsampled via a linear interpolation ($\textbf{x}{\downarrow}$) to match the specific temporal resolution of the intermediate decoder stages. The auxiliary weights are set to $\alpha_1 = 0.2 \text{ and } \alpha_2 = 0.3$ to prioritize the final high-resolution reconstruction while providing sufficient gradient guidance to the lower hierarchical layers.


\subsubsection{Fine-Tuning Design: The Spectral Mixture of Experts}
Once the SpecHi-Net backbones are pretrained, they are transitioned into a Spectral Mixture of Experts (SpecMoE) framework for downstream fine-tuning. This architecture enables the model to dynamically weight representations from multiple foundation models based on the spectral characteristics of the input signal.

\subsubsection{Fine-Tuning Pipeline}
The fine-tuning pipeline transforms a raw EEG segment $\textbf{x} \in \mathbb{R}^{C \times L}$ into a task-specific prediction through four primary stages: 
{\em (1) Parallel Expert Encoding:} the input $x$ is processed by three pretrained SpecHi-Net encoders (referred to as Experts $E_1$, $E_2$, and $E_3$) in parallel. Each expert produces a high-level embedding $O_i \in \mathbb{R}^{C \times D}$, where $C$ is the channel dimension and $D$ the embedding dimension. 
{\em (2) Spectral Gating:} simultaneously, the Power Spectral Density (PSD) of the raw input $x$ is computed to generate a gating signal. This signal determines the relevance of each expert's features based on the input's rhythmic content.
{\em (3) Gated Fusion:} the expert embeddings are modulated by their respective spectral gates and concatenated to form a unified representation $\textbf{out} \in \mathbb{R}^{C \times 3D}$. 
{\em (4) Hierarchical Pooling and Prediction:} the fused features undergo a pooling process on the spatial dimension before being passed to an MLP-based predictor for the final classification or regression task.

\subsubsection{Pretrained Experts}
We employ three distinct SpecHi-Net instances as our experts. 
Each expert develops complementary feature representations due to training on different data partitions with independent random masking seeds. 
During the initial phase of fine-tuning, these SpecHi-Net encoders can be either "frozen" to preserve the learned foundation features or "unfrozen" for task-specific adaptation. Each expert specializes in different aspects of the signal due to the stochastic nature of the pretraining masking process, providing the ensemble with a diverse set of perspectives on the neural data.

\subsubsection{Spectral Gating}
The core innovation of the fine-tuning framework is the {\em spectral gating mechanism}. Unlike traditional MoE models that use a learned router on latent features, SpecMoE utilizes the frequency information of the signal.
%
We compute the Power Spectral Density $\text{PSD}_x \in \mathbb{R}^{C \times K}$ (where $K$ is the frequency dimension) using a differentiable Welch's method. This provides a clear fingerprint of the signal's frequency distribution.
%
For each expert $E_i$, a gating tensor $G_i$ is generated through a linear transformation of $\text{PSD}_x$ followed by a sigmoid activation:
\begin{equation}
G_i = \sigma(\mathbf{W}_i \cdot \text{PSD}_x + b_i).
\end{equation}
These gates act as specific masks over the expert embeddings, amplifying or suppressing features based on whether the input signal's spectral profile matches the expert's learned "specialization".

\subsubsection{Predictor and Loss Functions}
Following gated fusion, the representation is regularized through a spatial pooling layer across the channel dimension $C$. This yields a compact vector $h \in \mathbb{R}^{3D}$ representing the global state of the EEG segment.
The final Predictor is a Multi-Layer Perceptron (MLP) with layer normalization and GELU activations. The framework is optimized based on the task: {\em (1) Classification}, optimized through a weighted cross-entropy loss to account for class imbalances when necessary. {\em (2) Regression}, optimized through Root Mean Squared Error (RMSE).

\section{Experimental design}
\subsubsection{Pretraining Dataset}
We pretrained SpecMoE on three subsets of the Temple University Hospital EEG dataset (TUEG) \cite{obeid2016temple}, comprising approximately 27,000 hours of clinical recordings from 14,987 subjects. To ensure data quality and consistency, we follow established preprocessing protocols for the EEG foundation models \cite{jianglarge,wangcbramod,zhou2025csbrain}, resulting in a curated set of 1,109,545 30-second samples ($\approx$9,000 hours). We partitioned this corpus into three sequential subsets based on subject identifiers to prevent data leakage and facilitate parallel expert training. Detailed preprocessing steps, including channel selection, filtering criteria, and bad-sample removal thresholds, are provided in the Supplementary Information (SI).

\subsubsection{Pretraining Implementation}
We trained each SpecHi-Net expert independently on an NVIDIA Tesla V100 4-GPU cluster. We utilized the AdamW optimizer with a cosine annealing learning rate scheduler to handle the large-scale self-supervised objective. 
Hyperparameter configurations, including learning rates, batch sizes, and hardware specifications, are detailed in the SI.


\subsubsection{Fine-Tuning Setup}
To comprehensively evaluate the generalizability of SpecMoE, we selected nine heterogeneous datasets representing distinct BCI paradigms and clinical diagnostics. These benchmarks cover a broad spectrum of neural activity, ranging from cognitive states and motor intent to drug-induced effects detection. 
A summary of the downstream tasks and their corresponding dataset characteristics is provided in Table \ref{tab:downstream_summary}. To maintain consistency with the pretraining phase, we resampled all EEG signals to 200 Hz. 
While we maintain a standardized backbone, specific artifact removal and segmenting procedures were tailored to the requirements of each task; detailed descriptions of these dataset-specific preprocessing pipelines are provided in the SI.

We benchmark SpecMoE against two categories of models: task-specific architectures and state-of-the-art foundation models. Among task-specific architectures, we selected three widely recognized baselines: EEGNet \cite{lawhern2018eegnet}, EEG-Conformer \cite{kasthuri2024eeg}, and FFCL \cite{li2022motor}. 
Regarding EEG foundation architectures, we selected three leading models: LaBraM \cite{jianglarge}, CBraMod \cite{wangcbramod}, and CSBrain \cite{zhou2025csbrain}. Detailed architectural differences are further discussed in the SI.

We quantified the performance on classification tasks using balanced accuracy, weighted F1-score, weighted AUROC, and weighted AUPRC. 
For the vigilance estimation regression task, we report the coefficient of determination ($R^2$), root mean square error (RMSE), and the Pearson correlation coefficient ($r$).

\begin{table}[t]
\centering
\caption{
Summary of downstream EEG datasets and task specifications. 
\textbf{Spec.} are the species (H: Human, M: Murine), 
\textbf{S.Freq} is the sampling frequency, 
\textbf{\#Ch.} is the number of channels, 
\textbf{Len} is the sample length, 
\textbf{\#Subj.} is the number of subjects.}
\label{tab:downstream_summary}
\setlength{\tabcolsep}{4pt} 
\resizebox{\textwidth}{!}{%
\begin{tabular}{@{}llccccccc@{}}
\toprule
\textbf{Task}          & \textbf{Dataset} & \textbf{Spec.} & \textbf{S.Freq} & \textbf{\#Ch.} & \textbf{Len} & \textbf{\#Samples} & \textbf{\#Subj.} & \textbf{Label} \\ \midrule
Motor Imagery          & PhysioNet-MI     & H              & 160 Hz          & 64             & 4s           & 9,837              & 109              & 4-class        \\
Emotion Recognition    & SEED-V           & H              & 1000 Hz         & 62             & 1s           & 117,744            & 16               & 5-class        \\
Sleep Staging          & HMC              & H              & 256 Hz          & 4              & 30s          & 137,243            & 151              & 4-class        \\
Therapeutic Area       & MACO             & M              & 1024 Hz         & 2              & 60s          & 61,900             & 336              & 5-class        \\
Drug Effect            & DA-Pharmaco      & M              & 1000 Hz         & 5              & 60s          & 2,800              & 10               & 5-class        \\
Imagined Speech        & BCIC2020-3       & H              & 256 Hz          & 64             & 3s           & 6,000              & 15               & 5-class        \\
Abnormal Detection     & TUAB             & H              & 256 Hz          & 16             & 10s          & 409,455            & 2,383            & 2-class        \\
Seizure Detection      & Siena            & H              & 512 Hz          & 29             & 10s          & 51,307             & 14               & 2-class        \\
Vigilance Estimation   & SEED-VIG         & H              & 200 Hz          & 17             & 8s           & 20,355             & 21               & continuous       \\ \bottomrule
\end{tabular}%
}
\end{table}


\section{Results}
Table \ref{tab:main_results} summarizes the performance of SpecMoE compared to the baselines.
For the sake of readability, we only report the primary metric for each task here. Detailed results can be found in the SI.
\begin{table}[ht!]
\centering
\caption{
Performance comparison on diverse EEG datasets. Balanced accuracy ($\uparrow$) is reported for classification tasks, and RMSE ($\downarrow$) is reported for SEED-VIG. Each value reported corresponds to the average across 5 random seeds, along with their standard deviations. \textbf{Bold} indicates the best performance; \underline{underline} indicates the second best.}
\label{tab:main_results}
\setlength{\tabcolsep}{2.2pt} 
\resizebox{\textwidth}{!}{
\begin{tabular}{@{}l ccccccc@{}}
\toprule
 & \multicolumn{3}{c}{\textbf{Task-Specific Models}} & \multicolumn{4}{c}{\textbf{Foundation Models}} \\
 \cmidrule(lr){2-4} \cmidrule(lr){5-8}
\textbf{Dataset} & \textbf{EEGNet} & \textbf{EEGConf} & \textbf{FFCL} & \textbf{LaBraM} & \textbf{CBramod} & \textbf{CSBrain} & \textbf{SpecMoE} \\ 
\midrule
\multirow{2}{6em}{PhysioNet-MI} & 0.5814 & 0.6049 & 0.5726 & 0.6173 & 0.6174 & \underline{0.6304} & \textbf{0.6444} \\
 & \scriptsize $\pm$0.0125 & \scriptsize $\pm$0.0104 & \scriptsize $\pm$0.0092 & \scriptsize $\pm$0.0122 & \scriptsize $\pm$0.0036 & \scriptsize $\pm$0.0090 & \scriptsize $\pm$0.0109 \\
\midrule
\multirow{2}{6em}{SEED-V} & 0.2961 & 0.3537 & 0.3641 & 0.3976 & \underline{0.4091} & \textbf{0.4197} & 0.4033 \\
 & \scriptsize $\pm$0.0102 & \scriptsize $\pm$0.0112 & \scriptsize $\pm$0.0092 & \scriptsize $\pm$0.0138 & \scriptsize $\pm$0.0097 & \scriptsize $\pm$0.0033 & \scriptsize $\pm$0.0084 \\
\midrule
\multirow{2}{6em}{HMC} & 0.6534 & 0.7149 & 0.4427 & 0.7277 & 0.7269 & \underline{0.7345} & \textbf{0.7479} \\
 & \scriptsize $\pm$0.0122 & \scriptsize $\pm$0.0086 & \scriptsize $\pm$0.0702 & \scriptsize $\pm$0.0101 & \scriptsize $\pm$0.0041 & \scriptsize $\pm$0.0047 & \scriptsize $\pm$0.0086 \\
\midrule
\multirow{2}{6em}{MACO} & 0.6659 & 0.6119 & 0.4325 & 0.625 & \underline{0.7645} & 0.6926 & \textbf{0.8527} \\
 & \scriptsize $\pm$0.0166 & \scriptsize $\pm$0.0217 & \scriptsize $\pm$0.0105 & \scriptsize $\pm$0.0272 & \scriptsize $\pm$0.0118 & \scriptsize $\pm$0.0117 & \scriptsize $\pm$0.0033 \\
 \midrule
\multirow{2}{6em}{DA-Pharmaco} & 0.4816 & \underline{0.5597} & 0.3074 & 0.4601 & 0.5344 & 0.4526 & \textbf{0.6229} \\
 & \scriptsize $\pm$0.0159 & \scriptsize $\pm$0.0345 & \scriptsize $\pm$0.0336 & \scriptsize $\pm$0.0487 & \scriptsize $\pm$0.021 & \scriptsize $\pm$0.0142 & \scriptsize $\pm$0.0231 \\
\midrule
\multirow{2}{6em}{BCIC2020-3} & 0.4413 & 0.4506 & 0.4678 & 0.5060 & 0.5373 & \underline{0.6004} & \textbf{0.6262} \\
 & \scriptsize $\pm$0.0096 & \scriptsize $\pm$0.0133 & \scriptsize $\pm$0.0197 & \scriptsize $\pm$0.0155 & \scriptsize $\pm$0.0108 & \scriptsize $\pm$0.0187 & \scriptsize $\pm$0.0166 \\
\midrule
\multirow{2}{6em}{TUAB} & 0.7642 & 0.7758 & 0.7848 & \underline{0.8140} & 0.7891 & \textbf{0.8172} & 0.7742 \\
 &\scriptsize $\pm$0.0036 & \scriptsize $\pm$0.0049 & \scriptsize $\pm$0.0038 & \scriptsize $\pm$0.0019 & \scriptsize $\pm$0.0030 & \scriptsize $\pm$0.0043 & \scriptsize $\pm$0.0069 \\
\midrule
\multirow{2}{6em}{SIENA} & 0.7487 & 0.7556 & 0.6616 & 0.7082 & 0.7317 & \underline{0.7662} & \textbf{0.8655} \\
 & \scriptsize $\pm$0.0521 & \scriptsize $\pm$0.0210 & \scriptsize $\pm$0.0391 & \scriptsize $\pm$0.0329 & \scriptsize $\pm$0.0647 & \scriptsize $\pm$0.0471 & \scriptsize $\pm$0.0038 \\
\midrule
\multirow{2}{6em}{SEED-VIG ($\downarrow$)} & 0.2847 & 0.2829 & 0.2885 & 0.2871 & 0.3057 & \underline{0.2774} & \textbf{0.1522} \\
 & \scriptsize $\pm$0.0076 & \scriptsize $\pm$0.0041 & \scriptsize $\pm$0.0093 & \scriptsize $\pm$0.0166 & \scriptsize $\pm$0.0027 & \scriptsize $\pm$0.0094 & \scriptsize $\pm$0.0029 \\
\bottomrule
\end{tabular}}
\end{table}
%
SpecMoE achieves {\em the best performance in 7 out of 9 downstream tasks}, demonstrating 
robustness across diverse recording conditions, including transfer to murine EEG data not seen during pretraining.
Our model exhibits its most significant gains in tasks characterized by complex spectral signatures and non-stationary dynamics.
Notably, on the MACO and SIENA datasets, our model maintains a substantial performance lead over the second-best foundation models (CBraMod and CSBrain) by 7.2\% and 9.9\%, respectively. In the MACO dataset, the challenge lies in distinguishing drug-induced modifications in murine brain activity across five classes (compounds from four therapeutic areas and solvents). In contrast, the SIENA dataset requires the identification of ictal (seizure) events against a background of interictal human EEG.
The consistent efficacy of SpecMoE in these two distinct benchmarks—one pharmacological and one pathological—indicates that our architecture framework is uniquely capable of prioritizing the specific expert features relevant to either the spectral signatures of pharmaceutical interventions or the high-energy discharges characteristic of seizures.

This competitive advantage is further highlighted in the DA-Pharmaco task, where SpecMoE achieves a balanced accuracy of 0.6229, surpassing the second-best model, EEGConformer, by 6\%. This task requires the classification of five distinct dopaminergic compounds from multi-site murine EEG  recordings. The success of SpecMoE here indicates that our architecture effectively decodes the complex neuro-spectral signatures and cross-regional dependencies induced by different dopaminergic modulators.
Furthermore, in the vigilance estimation regression task (SEED-VIG), SpecMoE nearly halves the error rate of competing models, achieving an RMSE of 0.1522. This improvement underscores the utility of the SpecMoE architecture in tracking subtle oscillatory shifts that characterize transitions between alertness and drowsiness.

A critical advantage of SpecMoE over recent transformer-based foundation models lies in its architectural flexibility. Many state-of-the-art models, such as CBraMod and CSBrain, rely on position-specific or channel-dependent embeddings that scale linearly with input dimensions. For datasets like MACO and DA-Pharmaco, which feature long sequence lengths (60 secs) or non-standard channel configurations, the parameter counts of these models escalate significantly—often exceeding 20 million parameters. More details on the parameters count can be found in the SI.
%
In contrast, SpecMoE utilizes 
RoPE and a hierarchical dual-path structure that are inherently invariant to sequence length and channel count. 
This allows SpecMoE to maintain a compact, standardized parameter footprint while delivering better accuracy, making it a more practical solution for real-world deployment across heterogeneous clinical environments and varying sensor montages.

\subsubsection{Ablation Studies}

\begin{figure}[t]
    \centering
    \includegraphics[width=1\linewidth]{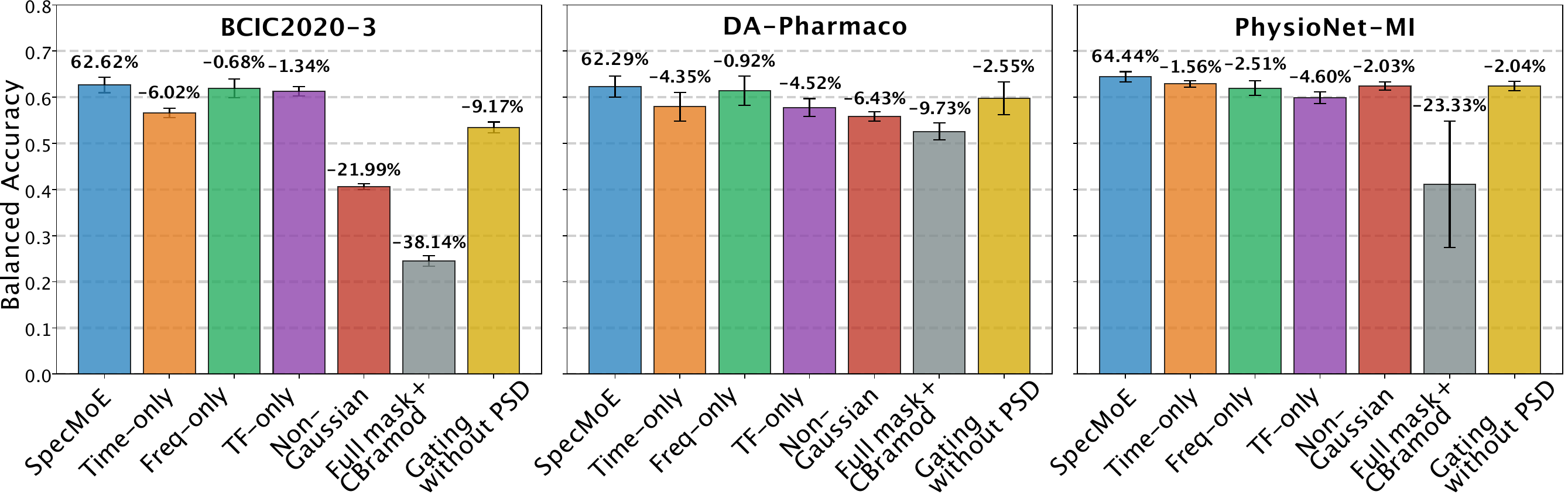}
    \caption{SpecMoE ablation results. We show the absolute value of the balanced accuracy for the SpecMoE model and relative differences for six ablation experiments. TF stands for time-frequency. See SI for other metrics.}
    \label{fig:fig3}
\end{figure}

To investigate the contribution of each component within the SpecMoE framework, we conducted a series of 6 ablation experiments across three representative datasets: BCIC2020-3  (imagined speech), DA-Pharmaco (pharmacology), and PhysioNet-MI (motor imagery). All ablation studies were performed using 3 random seeds.
Figure \ref{fig:fig3} 
illustrates the results.

A fundamental premise of our pretraining is that neural oscillations are best captured through a multi-dimensional corruption process. Our ablation results confirm this assumption. Indeed, all single-type masking strategies consistently underperform compared to the mixed-geometry approach. Training foundation models exclusively with time-only, frequency-only, or time-frequency-only masks results in a noticeable performance degradation across all benchmarks (see SI for the specific configuration of these single-type ablation experiments). For example, in the pharmacological task (Fig. \ref{fig:fig3} middle), the balanced accuracy drops from 0.6229 to 0.5794 (\textbf{Time-only}),  0.6137 (\textbf{Freq-only}), and 0.5777 (\textbf{TF-only}), indicating that a joint spectral-temporal objective is required to internalize the full complexity of EEG dynamics.
Furthermore, replacing our Gaussian-smoothed masks with traditional rectangular masks (\textbf{Non-Gaussian} experiment) leads to a significant drop in accuracy. 
For example, the balanced accuracy drops by more than 20\% in the imagined speech task (Fig. \ref{fig:fig3} left).

A critical finding of our study is the performance gap observed when replacing our SpecHi-Net backbone with the CBraMod architecture within our MoE framework (\textbf{Full mask+CBraMod}). We specifically selected CBraMod as a baseline because its backbone represents the current standard for high-performance, non-hierarchical EEG Transformers—an architecture it shares with CSBrain. By selecting CBraMod, we aimed to evaluate whether a competitive, flat transformer architecture could achieve performance parity with our model without the added complexity of the region-specific channel encoding layers found in CSBrain. Our goal was to isolate the impact of the hierarchical feature extraction process itself.
On the BCIC2020-3 dataset, this swap results in a catastrophic performance collapse from 0.6262 to 0.2448 (Fig. \ref{fig:fig3} left). 

Finally, we assess the benefit of using power spectral density (PSD) as a gating signal compared to a standard learned input encoding network (\textbf{Gating without PSD}). Detailed architecture and training parameters for this alternative learned gating network are provided in the SI. Our results show that utilizing a raw-data-driven encoding network consistently underperforms compared to our spectral gating strategy. 
The difference is particularly striking on the imagined speech task (Fig. \ref{fig:fig3} left).

\section{Conclusion}

In this work, we introduced \textbf{SpecMoE}, a spectral-anchored foundation model that utilizes a novel Gaussian-smoothed masking strategy to mitigate the "artificial transient" bias common in rectangular masking. By shifting the pretraining objective toward the reconstruction of smooth, endogenous neural rhythms, our framework develops a more physiologically grounded representation of EEG dynamics. Across nine heterogeneous datasets, SpecMoE achieves competitive performance, establishing a new baseline for several benchmarks. SpecMoE's success on both human and murine recordings highlights that the representations learned from human EEG transfer effectively to murine data after fine-tuning, suggesting shared spectral-temporal structure across species. Ablation studies indicate that the hierarchical \textbf{SpecHi-Net} architecture is instrumental for high-fidelity signal recovery in complex tasks, while the spectral-guided mixture-of-experts framework enables structured feature integration based on input rhythmic content. Together, these components provide a robust inductive bias for the next generation of universal EEG foundation models.

\subsubsection{Limitations}
Despite state-of-the-art results, our work has several limitations that provide avenues for future research. First, the requirement for full-parameter fine-tuning to achieve optimal performance on downstream tasks indicates that, while SpecMoE learns robust representations, a "zero-shot" universal EEG foundation model remains an open challenge. Second, while our mixed Gaussian masking strategy effectively mitigates boundary artifacts, we have not yet explored the impact of varying masking ratios; the current 50\% ratio selected following previous studies in EEG foundation models may be suboptimal.
Finally, the specialization of the experts within the MoE framework results from a stochastic data partitioning. Future work could investigate whether pretraining different experts using specific datasets from physiological domains such as sleep, pathology, or cognitive motor tasks would further enhance the efficiency of the spectral gating mechanism and therefore lead to better task predictions.
The current choice of three experts reflects a practical trade-off: the TUEG corpus naturally partitions into three subsets of sufficient size ($\sim$300k–400k samples each) to train robust individual models, while keeping the total computational cost tractable. 

\subsubsection{Ethical Considerations}
This work utilizes de-identified human and murine EEG data; all experimental protocols were approved by their respective Institutional Review Boards (IRB) or equivalent ethical committees. While SpecMoE is designed to advance clinical diagnostics and drug discovery, we acknowledge the inherent dual-use risks associated with high-capacity neural decoders, such as unauthorized cognitive surveillance. We strongly oppose any application of this technology that infringes upon cognitive liberty or is conducted without explicit informed consent. Furthermore, we emphasize that all automated assessments must undergo professional validation to mitigate risks from algorithmic bias and ensure patient safety across diverse populations.

\subsubsection{Acknowledgment}
This work was funded by SynapCell SAS through the Cortex project, awarded at the 9th edition of the i-Nov competition organized for French companies. This work was also granted access to the HPC resources of IDRIS under the allocation 2025-AD011016062R1 made by GENCI and the HPC ressources from GRICAD infrastructure which is supported by Grenoble research communities.

%


\newpage
\appendix
\title{Supplementary Information for SpecMoE: Spectral Mixture-of-Experts Foundation Model for Cross-Species EEG Decoding}

\setcounter{section}{0}
\setcounter{table}{0}
\setcounter{figure}{0}
\renewcommand{\thesection}{\Alph{section}}
\renewcommand{\thetable}{S\arabic{table}}
\renewcommand{\thefigure}{S\arabic{figure}}



\author{Author information scrubbed for double-blind reviewing}
\section*{\centering Supplementary Information } 

\setcounter{section}{0}
\setcounter{table}{0}
\setcounter{figure}{0}
\renewcommand{\thesection}{\Alph{section}}
\renewcommand{\thetable}{S\arabic{table}}
\renewcommand{\thefigure}{S\arabic{figure}}

\section{Pretraining Setup}
\subsection{Pretraining Dataset and Preprocessing}
\subsubsection{Dataset Description}
We pretrained SpecMoE on 3 subsets of the large Temple University Hospital EEG dataset \cite{obeid2016temple} (TUEG). The TUEG dataset comprises a diverse archive of 69,652 clinical EEG recordings from 14,987 subjects across 26,846 sessions, totaling 27,062 hours in duration. The archive features over 40 different channel configurations and recordings of varying durations. Most of the recordings are sampled at 256 Hz. Unfortunately, the TUEG dataset suffers from significant data contamination, including a substantial amount of unmarked noise, artifacts, and faulty channels. To homogenize the pretraining process and reduce the dataset inconsistencies, we preprocessed it following previous studies on EEG foundation models. These include LaBraM \cite{jianglarge}, CBraMod \cite{wangcbramod}, and CSBrain \cite{zhou2025csbrain}.

\subsubsection{Preprocessing}

To begin with, we remove recordings that have a total duration of less than 5 minutes. Next, we discard the first and last minute of each recording to eliminate as much low-quality data as possible. We then select 19 common EEG channels (Fp1, Fp2, F7, F3, Fz, F4, F8, T3, C3, Cz, C4, T4, T5, P3, Pz, P4, T6, O1, O2) that comply with a subset of the 10-20 international electrode placement system standards. This ensures that we obtain clean and uniformly formatted pre-training data. 

Afterward, we apply a band-pass filter (0.3 Hz–75 Hz) to eliminate low-frequency and high-frequency noise. Additionally, a notch filter (60 Hz) is utilized to remove power line noise. All EEG signals are resampled to 200 Hz and segmented into non-overlapping 30-second EEG samples. 

However, the preprocessing steps mentioned above may not completely resolve the quality issues present in the EEG data. To further enhance the quality, we implement an automated scheme for the removal of bad EEG samples. Specifically, we identify samples as bad if any data point exceeds an absolute amplitude of 100 $\mu$V and remove these from the dataset. We also normalize the EEG signals by scaling the units to 100 $\mu$V, ensuring that the values predominantly range between -1 and 1, in line with previous work on EEG foundation models (LaBraM \cite{jianglarge}, CBraMod \cite{wangcbramod}, CSBrain\cite{zhou2025csbrain}). After removing bad samples, we are left with 1,109,545 remaining 30-second segments, totaling 9,000 hours of data.

Finally, we divide the preprocessed samples into three subsets. Subset-1 contains samples 1 to 400,000, subset-2 consists of samples 400,001 to 800,000, and subset-3 includes the remaining samples from 800,001 to 1,109,545. This sequential splitting strategy is based on the hierarchical storage structure of the corpus, where patients are organized into folders according to randomized database identifiers. This ensures that each subset represents a diverse and unbiased cross-section of the 14,987 unique subjects. Furthermore, considering the large number of samples per subset and the relatively low average number of sessions per patient (1.79), this approach effectively minimizes the risk of data leakage between the three training partitions.

\subsubsection{Computational Details and Hyperparameters}
SpecMoE was implemented using the PyTorch framework (v2.6.0) and trained in a high-performance computing environment. To develop the three specialized experts within our foundation ensemble, each model was trained independently on one of the three TUEG subsets described in Section 4.1.2. 
The hardware configuration consisted of 4 NVIDIA Tesla V100-SXM2-32GB GPUs operating in parallel. Each pretraining session spanned approximately 200 hours to complete 50 epochs. We utilized the AdamW optimizer with an initial learning rate of $1 \times 10^{-3}$ and a weight decay coefficient of $5 \times 10^{-2}$. To ensure stable convergence, we employed a \texttt{CosineAnnealingLR} scheduler. 
The effective batch size was set to 256 samples, achieved through a distributed strategy of 64 samples across the 4  GPUs with 4 gradient accumulation steps. Signal processing and data management were supported by NumPy (v2.3.3) and SciPy (v1.16.2) within a Python 3.12.8 environment. All computations were executed using CUDA 12.4 to leverage hardware acceleration.
%
Table \ref{tab:pretraining_hyperparams} lists the hyperparameters used for the pretraining phase.

\begin{table}[!ht]
    \centering
    \caption{Hyperparameters for SpecMoE pre-training.}
    \label{tab:pretraining_hyperparams}
    \setlength{\tabcolsep}{6pt}
    \begin{tabular}{ll}
        \toprule
        \textbf{Category} & \textbf{Setting} \\ \midrule
        \textbf{Input EEG Data} & \\
        Channels ($C$) & 19 \\
        Sampling Frequency & 200 Hz \\
        Sample Duration ($L$) & 30s (6000 points) \\
        Amplitude Normalization & Unit 100 $\mu$V \\ \midrule
        \textbf{Signal Transformation (STFT)} & \\
        Window Type & hann\_periodic \\
        Window Length ($n\_fft$) & 400 \\
        Hop Length ($n\_overlap$) & 200 \\
        Frequency Bins ($F$) & 201 \\
        Time Frames ($T$) & 31 \\ \midrule
        \textbf{Gaussian Masking} & \\
        Total Mask Ratio ($\rho$) & 0.5 \\
        Geometries Probability ($P$) & [Freq: 0.6, Time: 0.3, TF: 0.1] \\
        Kernel parameters ($\sigma_f, \sigma_t$) & 5\% of respective range \\
        Spectral-Band Bias & 50\% on $\delta, \theta, \alpha, \beta$ bands \\ \midrule
        \textbf{SpecHi-Net Backbone} & \\
        Hierarchy Stages & 3 Downsampling / 3 Upsampling \\
        Dual-Path Kernels ($k$) & Small: 4, Large: 65 \\
        Transformer Layers & 1 \\
        Attention Heads & 8 \\
        Hidden Dimension ($D$) & 128 \\
        Position Encoding & Rotary Positional Embedding (RoPE) \\ \midrule
        \textbf{Optimization} & \\
        Loss Function & Multi-Objective (MSE + Spectral Loss) \\
        Spectral Loss Weight ($w_{spec}$) & 0.02 \\
        Auxiliary Loss Weights ($\alpha_1, \alpha_2$) & 0.2, 0.3 \\
        Optimizer & AdamW \\
        Learning Rate & $1 \times 10^{-3}$ \\
        Weight Decay & $5 \times 10^{-2}$ \\
        Scheduler & CosineAnnealingLR \\
        Batch Size & 256 (64 per GPU $\times$ 4 steps) \\
        Epochs & 50 \\
        Gradient Clipping & 5.0 \\ \bottomrule
    \end{tabular}
\end{table}

\section{Details on Downstream Tasks}

\subsection{Finetuning Settings Details}
\subsubsection{Baselines}
\begin{itemize}
    \item \textbf{EEGNet:} A compact, specialized CNN for EEG that utilizes depthwise and separable convolutions to extract robust features with a minimal parameter footprint \cite{lawhern2018eegnet}. Code available at: \url{https://github.com/vlawhern/arl-eegmodels}.
    
    \item \textbf{EEGConformer:} A hybrid architecture that leverages CNNs for local feature extraction and Transformer blocks to capture long-range temporal dependencies \cite{kasthuri2024eeg}. Code available at: \url{https://github.com/eeyhsong/EEG-Conformer}.
    
    \item \textbf{FFCL:} A framework integrating parallel CNN and LSTM branches to fuse spatial and temporal dynamics through a shared fully connected layer \cite{li2022motor}. We implemented the code ourselves, following the architecture design and the detailed hyperparameter description in the paper.
    
    \item \textbf{LaBraM:} A unified foundation model that enables cross-dataset learning through neural channel patching and a vector-quantized neural spectrum tokenizer. It utilizes a Transformer backbone and dual-domain (frequency/phase) mask learning for large-scale pretraining \cite{jianglarge}. Code available at: \url{https://github.com/935963004/LaBraM/tree/main}.
    
    \item \textbf{CBraMod:} A foundation model featuring criss-cross attention to capture spatial and temporal features separately, achieving efficient yet powerful representations \cite{wangcbramod}. Code available at: \url{https://github.com/wjq-learning/CBraMod/tree/main}.
    
    \item \textbf{CSBrain:} A model employing cross-scale spatiotemporal tokenization and structured sparse attention to optimize robustness in EEG signal decoding \cite{zhou2025csbrain}. Code available at: \url{https://github.com/yuchen2199/CSBrain/tree/main}.
\end{itemize}

\subsubsection{Metrics}
\paragraph{Classification Metrics.}
For a given class $c \in \{1, \dots, C\}$, let $TP_c$, $TN_c$, $FP_c$, and $FN_c$ represent the True Positives, True Negatives, False Positives, and False Negatives, respectively. We define the fundamental components for each class as:
\begin{equation}
\text{Precision}_c (P_c) = \frac{TP_c}{TP_c + FP_c}, \quad \text{Recall}_c (R_c) = \frac{TP_c}{TP_c + FN_c}.
\end{equation}
Additionally, for threshold-based curves, we define the True Positive Rate ($\text{TPR} = \text{Recall}$) and the False Positive Rate ($\text{FPR} = \frac{FP}{FP+TN}$). Based on these, the following metrics are employed:
\begin{itemize}
    \item \textbf{Balanced Accuracy:} The arithmetic mean of class-specific Recall scores, which prevents inflated performance estimates on imbalanced datasets:
    \begin{equation}
    \text{Balanced Acc.} = \frac{1}{C} \sum_{c=1}^{C} R_c.
    \end{equation}

    \item \textbf{Weighted F1-Score:} The harmonic mean of Precision and Recall, weighted by the number of samples ($N_c$) in each class relative to the total samples ($N_{total}$):
    \begin{equation}
        \text{Weighted F1} = \sum_{c=1}^{C} \frac{N_c}{N_{total}} \left( 2 \cdot \frac{P_c \cdot R_c}{P_c + R_c} \right).
    \end{equation}
    
    \item \textbf{AUROC:} The area under the curve obtained by plotting TPR against FPR at various decision thresholds:
    \begin{equation}
        \text{AUROC} = \int_{0}^{1} \text{TPR}(f) \, df, \quad f = \text{FPR}.
    \end{equation}
    
    \item \textbf{AUPRC:} The area under the curve obtained by plotting Precision against Recall, providing a stringent assessment for minority class detection:
    \begin{equation}
        \text{AUPRC} = \int_{0}^{1} P(r) \, dr, \quad r = \text{Recall}.
    \end{equation}
\end{itemize}

\paragraph{Regression Metrics (Vigilance Estimation).}For the SEED-VIG dataset, let $y_i$ be the ground truth, $\hat{y}_i$ the predicted value, and $\bar{y}$ the mean of the ground truth for $n$ samples.
\begin{itemize}
    \item \textbf{Pearson Correlation Coefficient ($r$):} Measures the linear relationship between predicted and true labels:
\begin{equation}r = \frac{\sum_{i=1}^{n} (y_i - \bar{y})(\hat{y}i - \bar{\hat{y}})}{\sqrt{\sum{i=1}^{n} (y_i - \bar{y})^2 \sum_{i=1}^{n} (\hat{y}_i - \bar{\hat{y}})^2}}.
\end{equation}

    \item \textbf{Coefficient of Determination ($R^2$):} Represents the proportion of variance explained by the model:
    \begin{equation}
        R^2 = 1 - \frac{\sum_{i=1}^{n} (y_i - \hat{y}_i)^2}{\sum_{i=1}^{n} (y_i - \bar{y})^2}.
    \end{equation}
    
    \item \textbf{Root Mean Square Error (RMSE):} Quantifies the standard deviation of the prediction residuals:
    \begin{equation}
        \text{RMSE} = \sqrt{\frac{1}{n} \sum_{i=1}^{n} (y_i - \hat{y}_i)^2}.
    \end{equation}
\end{itemize}

\subsubsection{Hyperparameters}
In Table \ref{tab:finetuning_hyperparams}, we specify the default hyperparameter setup used for the finetuning tasks. Details on each dataset can be found in the code: \url{https://github.com/TeraXj78/SpecMoE}.
\begin{table}[!ht]
    \centering
    \small
    \caption{Hyperparameters for SpecMoE downstream finetuning.}
    \label{tab:finetuning_hyperparams}
    \setlength{\tabcolsep}{6pt}
    \begin{tabular}{ll}
        \toprule
        \textbf{Category} & \textbf{Setting} \\ \midrule
        \textbf{Architecture Setup} & \\
        Number of Experts & 3 (Pretrained SpecHi-Net encoders) \\
        Expert Backbone State & Unfrozen (Full-parameter finetuning) \\
        Embedding Dimension ($D$) & 128 (per expert) \\
        Fused Representation Dim & 384 ($3 \times D$) \\ \midrule
        \textbf{Spectral Gating (PSD)} & \\
        PSD Method & Differentiable Welch's method \\
        Gating Input Dim & 201 (Frequency Bins $F$) \\
        Gating Activation & Sigmoid \\ \midrule
        \textbf{Downstream Predictor} & \\
        Spatial Pooling & Average pooling across channels \\
        MLP Layers & 2 (Linear $\rightarrow$ GELU $\rightarrow$ Linear) \\
        Dropout Rate & 0.2 \\ \midrule
        \textbf{Optimization} & \\
        Loss Function (Classification) & Weighted Cross-Entropy \\
        Loss Function (Regression) & Root Mean Square Error (RMSE) \\
        Optimizer & Adam \\
        Learning Rate & $1 \times 10^{-3}$ \\
        Weight Decay & $0$ \\
        Scheduler & ReduceLROnPlateau \\
        Batch Size & 64 \\
        Max Epochs & 100 \\ \bottomrule
    \end{tabular}
\end{table}

\subsection{Motor Imagery Task}
The PhysioNet-MI dataset \cite{schalk2004bci2000} serves as a large-scale benchmark for decoding motor intent, featuring 109 subjects performing four mental imagery tasks: left/right fist, both fists, and both feet. Data from 64 channels were resampled to 200 Hz and windowed into 4-second trials, totaling 9,837 samples. We evaluate the model using a strict subject-independent protocol, allocating subjects 1--70 for training, 71--89 for validation, and 90--109 for testing.

As detailed in Table \ref{tab:results_physionet_mi}, foundation models demonstrate a clear advantage over task-specific baselines in generalizing across a diverse population. SpecMoE achieves the highest performance with a balanced accuracy of \textbf{0.6444} and a weighted F1-score of \textbf{0.6476}. Notably, SpecMoE surpasses the strongest foundation model baseline, CSBrain, by \textbf{1.40\%} in balanced accuracy and \textbf{1.68\%} in weighted F1. Compared to the best-performing task-specific model (EEGConformer), our framework yields a substantial improvement of \textbf{3.95\%} in accuracy. The results suggest that our spectral mixture-of-experts is effective at capturing the rhythmic signatures necessary for high-fidelity motor imagery classification across unseen subjects.

\begin{table}[ht]
\centering
\caption{Detailed performance comparison on the \textbf{PhysioNet-MI} motor imagery task. Values represent the average and standard deviation across five random seeds. \textbf{Bold} indicates the best performance; \underline{underline} indicates the second best.}
\label{tab:results_physionet_mi}
\setlength{\tabcolsep}{3pt}
\resizebox{\textwidth}{!}{%
\begin{tabular}{@{}llccccccc@{}}
\toprule
 & & \multicolumn{3}{c}{\textbf{Task-Specific Models}} & \multicolumn{4}{c}{\textbf{Foundation Models}} \\
 \cmidrule(lr){3-5} \cmidrule(lr){6-9}
\textbf{} & \textbf{Sub-metric} & \textbf{EEGNet} & \textbf{EEGConf} & \textbf{FFCL} & \textbf{LaBraM} & \textbf{CBramod} & \textbf{CSBrain} & \textbf{SpecMoE} \\ \midrule
\textbf{\# Params} & --- & --- & --- & --- & 5.8 M & 34.2 M & 38.2 M & 4.3 M \\ \midrule
\multirow{2}{*}{\textbf{Bal. Acc}} & Mean & 0.5814 & 0.6049 & 0.5726 & 0.6173 & 0.6174 & \underline{0.6304} & \textbf{0.6444} \\
 & Std Dev & $\pm$0.0125 & $\pm$0.0104 & $\pm$0.0092 & $\pm$0.0122 & $\pm$0.0036 & $\pm$0.0090 & $\pm$0.0109 \\ \midrule
\multirow{2}{*}{\textbf{Weighted F1}} & Mean & 0.5796 & 0.6062 & 0.5701 & 0.6177 & 0.6179 & \underline{0.6308} & \textbf{0.6476} \\
 & Std Dev & $\pm$0.0115 & $\pm$0.0095 & $\pm$0.0079 & $\pm$0.0141 & $\pm$0.0035 & $\pm$0.0095 & $\pm$0.0082 \\ \bottomrule
\end{tabular}%
}
\end{table}

\subsection{Emotion Recognition Task}
SEED-V \cite{liu2021comparing} is a benchmark dataset for EEG-based emotion recognition, featuring five emotional categories: happy, sad, neutral, disgust, and fear. EEG signals were collected from 16 subjects across three sessions using 62 channels at a sampling rate of 1000 Hz. Following standard preprocessing, the data was resampled to 200 Hz and segmented into 1-second windows, resulting in 117,744 samples. We employ a within-session split protocol, dividing the 15 trials of each session into three equal parts (5:5:5) for training, validation, and testing.

As shown in Table \ref{tab:results_seed_v}, decoding high-granularity emotional states remains a significant challenge for all architectures, with performance across the board generally lower than on simpler motor or clinical tasks. While SpecMoE does not achieve the top rank in this specific paradigm, it remains highly competitive with a balanced accuracy of \textbf{0.4033} and a weighted F1-score of \textbf{0.4142}. Specifically, SpecMoE performs within a \textbf{1.64\%} margin of the state-of-the-art CSBrain and continues to outperform earlier foundation models like LaBraM, as well as all task-specific baselines. These results suggest that while the spectral gating mechanism is effective, the extremely short 1-second temporal context of SEED-V may limit the full potential of hierarchical spectral modeling compared to datasets with longer rhythmic signatures.

\begin{table}[ht]
\centering
\caption{Detailed performance comparison on the \textbf{SEED-V} emotion recognition task. Values represent the average and standard deviation across five random seeds. \textbf{Bold} indicates the best performance; \underline{underline} indicates the second best.}
\label{tab:results_seed_v}
\setlength{\tabcolsep}{3pt}
\resizebox{\textwidth}{!}{%
\begin{tabular}{@{}llccccccc@{}}
\toprule
 & & \multicolumn{3}{c}{\textbf{Task-Specific Models}} & \multicolumn{4}{c}{\textbf{Foundation Models}} \\
 \cmidrule(lr){3-5} \cmidrule(lr){6-9}
\textbf{} & \textbf{Sub-metric} & \textbf{EEGNet} & \textbf{EEGConf} & \textbf{FFCL} & \textbf{LaBraM} & \textbf{CBramod} & \textbf{CSBrain} & \textbf{SpecMoE} \\ \midrule
\textbf{\# Params} & --- & --- & --- & --- & 5.8 M & 14.3 M & 18.3 M & 4.3 M \\ \midrule
\multirow{2}{*}{\textbf{Bal. Acc}} & Mean & 0.2961 & 0.3537 & 0.3641 & 0.3976 & \underline{0.4091} & \textbf{0.4197} & 0.4033 \\
 & Std Dev & $\pm$0.0102 & $\pm$0.0112 & $\pm$0.0092 & $\pm$0.0138 & $\pm$0.0097 & $\pm$0.0033 & $\pm$0.0084 \\ \midrule
\multirow{2}{*}{\textbf{Weighted F1}} & Mean & 0.2749 & 0.3487 & 0.3645 & 0.3974 & \underline{0.4101} & \textbf{0.4280} & 0.4142 \\
 & Std Dev & $\pm$0.0098 & $\pm$0.0136 & $\pm$0.0132 & $\pm$0.0111 & $\pm$0.0108 & $\pm$0.0023 & $\pm$0.0091 \\ \bottomrule
\end{tabular}%
}
\end{table}

\subsection{Sleep Stages Classification Task}
The HMC (Haaglanden Medisch Centrum) dataset \cite{PhysioNet-hmc-sleep-staging-1.1} is a standard benchmark for sleep stage scoring. It contains polysomnography (PSG) recordings from 151 subjects, from which we extract the EEG signals from four channels (F4-M1, C4-M1, O2-M1, and C3-M2). The task involves classifying 30-second epochs into five stages defined by the American Academy of Sleep Medicine manual: Wake (W), Non-REM 1 (N1), Non-REM 2 (N2), Non-REM 3 (N3), and REM (R). The signals were resampled to 200 Hz, consistent with our pretraining configuration. For this task, we followed the subject-independent split used in recent literature \cite{zhou2025csbrain}.

As shown in Table \ref{tab:results_hmc}, SpecMoE achieves the highest balanced accuracy of \textbf{0.7479}, surpassing the best foundation baseline, CSBrain (\textbf{0.7345}), by \textbf{1.34\%}. In terms of the weighted F1-score, SpecMoE (\textbf{0.7503}) remains highly competitive, performing on par with CSBrain (\textbf{0.7506}) within a negligible margin. Notably, all foundation models demonstrate a massive performance leap over the FFCL baseline, which struggled with the temporal complexity of sleep stages. The success of SpecMoE here is particularly meaningful; sleep stages are characterized by specific spectral landmarks like slow-wave delta activity, which our spectrally-anchored mixture-of-experts is designed to isolate and prioritize during the learning process.

\begin{table}[ht]
\centering
\caption{Detailed performance comparison on the \textbf{HMC} sleep stages classification task. Values represent the average and standard deviation across five random seeds. \textbf{Bold} indicates the best performance; \underline{underline} indicates the second best.}
\label{tab:results_hmc}
\setlength{\tabcolsep}{3pt}
\resizebox{\textwidth}{!}{%
\begin{tabular}{@{}llccccccc@{}}
\toprule
 & & \multicolumn{3}{c}{\textbf{Task-Specific Models}} & \multicolumn{4}{c}{\textbf{Foundation Models}} \\
 \cmidrule(lr){3-5} \cmidrule(lr){6-9}
\textbf{} & \textbf{Sub-metric} & \textbf{EEGNet} & \textbf{EEGConf} & \textbf{FFCL} & \textbf{LaBraM} & \textbf{CBramod} & \textbf{CSBrain} & \textbf{SpecMoE} \\ \midrule
\textbf{\# Params} & --- & --- & --- & --- & 5.8 M & 20.3 M & 24.3 M & 4.3 M \\ \midrule
\multirow{2}{*}{\textbf{Bal. Acc}} & Mean & 0.6534 & 0.7149 & 0.4427 & 0.7277 & 0.7269 & \underline{0.7345} & \textbf{0.7479} \\
 & Std Dev & $\pm$0.0122 & $\pm$0.0086 & $\pm$0.0702 & $\pm$0.0101 & $\pm$0.0041 & $\pm$0.0047 & $\pm$0.0086 \\ \midrule
\multirow{2}{*}{\textbf{Weighted F1}} & Mean & 0.6536 & 0.7080 & 0.2902 & 0.7454 & 0.7395 & \textbf{0.7506} & \underline{0.7503} \\
 & Std Dev & $\pm$0.0168 & $\pm$0.0039 & $\pm$0.0485 & $\pm$0.0027 & $\pm$0.0089 & $\pm$0.0042 & $\pm$0.0024 \\ \bottomrule
\end{tabular}%
}
\end{table}

\subsection{Drug Therapeutic Area Classification Task}
The MACO dataset is a large-scale, private repository designed to investigate the therapeutic potential of various pharmacological compounds through EEG signatures. It comprises $\sim$1,032 hours of recordings from 336 mice, covering four major therapeutic classes: antidepressants, antipsychotics, antiepileptics, and anxiolytics, alongside a fifth control group receiving solvent administration. The EEG was recorded from two brain regions—the prefrontal and parietal cortices, using 2-channel setups at a sampling rate of 1024 Hz. Following our standard pipeline, signals were resampled to 200 Hz and segmented into 60-second windows, totaling 61,900 samples. We employed a strict subject-independent split, allocating 70\% of the mice for training, 10\% for validation, and 20\% for testing.

As shown in Table \ref{tab:results_maco}, SpecMoE achieves its strongest performance on this benchmark, reaching a balanced accuracy of \textbf{0.8527} and a weighted F1-score of \textbf{0.8499}. This represents an absolute improvement of \textbf{8.82\%} in accuracy over the second-best model, CBraMod (\textbf{0.7645}), and a \textbf{16.01\%} lead over CSBrain. Notably, these results were achieved with a significantly more efficient parameter footprint; while models like CBraMod and CSBrain utilize \textbf{more than 20 million parameters}, SpecMoE maintains high-fidelity decoding with approximately \textbf{4.3 million parameters}.

The success on the MACO dataset is highly significant for two reasons. First, it demonstrates that our spectral-anchored pretraining, which was primarily conducted on human data, generalizes effectively to murine EEG, validating the \textbf{cross-species} utility of the foundation model. Second, the substantial performance gap suggests that therapeutic drug effects are deeply embedded in certain frequency-domain bands that our spectral mixture-of-experts model was specifically designed to capture.
These results indicate that SpecMoE is a powerful tool for pharmacological research and drug discovery, outperforming much larger models in identifying the neural signatures of therapeutic compounds.

\begin{table}[ht]
\centering
\caption{Detailed performance comparison on the \textbf{MACO} drug therapeutic area classification task. Values represent the average and standard deviation across five random seeds. \textbf{Bold} indicates the best performance; \underline{underline} indicates the second best.}
\label{tab:results_maco}
\setlength{\tabcolsep}{3pt}
\resizebox{\textwidth}{!}{%
\begin{tabular}{@{}llccccccc@{}}
\toprule
 & & \multicolumn{3}{c}{\textbf{Task-Specific Models}} & \multicolumn{4}{c}{\textbf{Foundation Models}} \\
 \cmidrule(lr){3-5} \cmidrule(lr){6-9}
\textbf{} & \textbf{Sub-metric} & \textbf{EEGNet} & \textbf{EEGConf} & \textbf{FFCL} & \textbf{LaBraM} & \textbf{CBramod} & \textbf{CSBrain} & \textbf{SpecMoE} \\ \midrule
\textbf{\# Params} & --- & --- & --- & --- & 5.8 M & 20.3 M & 24.3 M & 4.3 M \\ \midrule
\multirow{2}{*}{\textbf{Bal. Acc}} & Mean & 0.6659 & 0.6119 & 0.4325 & 0.6250 & \underline{0.7645} & 0.6926 & \textbf{0.8527} \\
 & Std Dev & $\pm$0.0166 & $\pm$0.0218 & $\pm$0.0105 & $\pm$0.0272 & $\pm$0.0119 & $\pm$0.0117 & $\pm$0.0033 \\ \midrule
\multirow{2}{*}{\textbf{Weighted F1}} & Mean & 0.6828 & 0.6585 & 0.4190 & 0.6736 & \underline{0.8013} & 0.7292 & \textbf{0.8499} \\
 & Std Dev & $\pm$0.0182 & $\pm$0.0192 & $\pm$0.0935 & $\pm$0.0388 & $\pm$0.0044 & $\pm$0.0189 & $\pm$0.0083 \\ \bottomrule
\end{tabular}%
}
\end{table}

\subsection{Drug Effects Classification Task}
The DA-Pharmaco dataset \cite{kapanaiah2024neural} is a specialized pharmacological benchmark consisting of Local Field Potential (LFP) recordings from depth electrodes in ten mice. Electrodes were implanted across five key regions: the prelimbic cortex (PrL), mediodorsal thalamus (MD), dorsal hippocampal fissure (dCA1), dorsal hippocampal CA3 subfield (dCA3), and ventral hippocampal fissure (vHC). The original study followed a within-subject, randomized Latin-squares design testing seven conditions: Saline, TWEEN80/saline, two doses of Clozapine (1 and 3 mg/kg), Raclopride, SCH23390, and Amphetamine.
To evaluate the model's capacity to decode fundamental pharmacological effects, we designed a specific 5-class grouping strategy: (1) Vehicles (Saline and TWEEN80), (2) Amphetamine, (3) Clozapine (grouping both 1 and 3 mg/kg doses), (4) SCH23390, and (5) Raclopride. We extracted a 40-minute window starting 5 minutes post-injection for each session. Signals were resampled to 200 Hz and segmented into 60-second samples (2,800 in total). We utilized a subject-independent split, assigning six subjects for training, two for validation, and two for testing.

As detailed in Table \ref{tab:results_da_pharmaco}, SpecMoE demonstrates a significant performance advantage, achieving a balanced accuracy of \textbf{0.6230} and a weighted F1-score of \textbf{0.6329}. This represents an improvement of \textbf{6.33\%} over the best task-specific model (EEGConformer) and a \textbf{8.86\%} lead over CBraMod. Notably, while larger foundation models like CSBrain (0.4526) and LaBraM (0.4601) struggled to distinguish these self-designed pharmacological classes, SpecMoE's spectral-anchored gating effectively isolated the regional oscillatory signatures associated with each drug category. These results suggest that SpecMoE is sensitive to the fine-grained spectral shifts that define specific neurotransmitter modulations, even in complex depth-electrode recordings.

\begin{table}[ht]
\centering
\caption{Detailed performance comparison on the \textbf{DA-Pharmaco} drug effects classification task. Values represent the average and standard deviation across five random seeds. \textbf{Bold} indicates the best performance; \underline{underline} indicates the second best.}
\label{tab:results_da_pharmaco}
\setlength{\tabcolsep}{3pt}
\resizebox{\textwidth}{!}{%
\begin{tabular}{@{}llccccccc@{}}
\toprule
 & & \multicolumn{3}{c}{\textbf{Task-Specific Models}} & \multicolumn{4}{c}{\textbf{Foundation Models}} \\
 \cmidrule(lr){3-5} \cmidrule(lr){6-9}
\textbf{} & \textbf{Sub-metric} & \textbf{EEGNet} & \textbf{EEGConf} & \textbf{FFCL} & \textbf{LaBraM} & \textbf{CBramod} & \textbf{CSBrain} & \textbf{SpecMoE} \\ \midrule
\textbf{\# Params} & --- & --- & --- & --- & 5.8 M & 38.7 M & 42.7 M & 4.3 M \\ \midrule
\multirow{2}{*}{\textbf{Bal. Acc}} & Mean & 0.4816 & \underline{0.5597} & 0.3074 & 0.4601 & 0.5344 & 0.4526 & \textbf{0.6230} \\
 & Std Dev & $\pm$0.0159 & $\pm$0.0345 & $\pm$0.0336 & $\pm$0.0487 & $\pm$0.0210 & $\pm$0.0142 & $\pm$0.0231 \\ \midrule
\multirow{2}{*}{\textbf{Weighted F1}} & Mean & 0.4658 & 0.5309 & 0.3080 & 0.4583 & \underline{0.5389} & 0.4575 & \textbf{0.6329} \\
 & Std Dev & $\pm$0.0127 & $\pm$0.0520 & $\pm$0.0403 & $\pm$0.0710 & $\pm$0.0288 & $\pm$0.0155 & $\pm$0.0206 \\ \bottomrule
\end{tabular}%
}
\end{table}

\subsection{Imagined Speech Recognition Task}
Imagined speech classification aims to decode phonological representations embedded in neural activity without any physical speech. This task is critical for developing augmentative communication technologies for individuals with severe speech impairments resulting from stroke or amyotrophic lateral sclerosis (ALS). We evaluate our model on the BCIC2020-3 dataset \cite{jeong20222020}, released for the 2020 International BCI Competition. In this experiment, 15 subjects imagined five speech-related categories ("hello", "help me", "stop", "thank you", and "yes"). EEG signals were recorded from 64 channels at 256 Hz and resampled to 200 Hz for our experiments. Following the competition's rigorous split, each subject provided 60 trials per class for training, 10 for validation, and 10 for testing, with each sample consisting of a 3-second recording.

As shown in Table \ref{tab:results_bcic_2020_3}, imagined speech recognition represents a high-complexity decoding task where task-specific models often struggle to exceed 50\% accuracy. SpecMoE achieves the highest performance across all models, with a balanced accuracy of \textbf{0.6262} and a weighted F1-score of \textbf{0.6264}. This represents a substantial improvement of \textbf{2.58\%} over the previous best-performing foundation model, CSBrain (\textbf{0.6004}). Furthermore, SpecMoE surpasses the strongest task-specific baseline (FFCL) by \textbf{15.84\%}, demonstrating that the spectral mixture-of-experts architecture is capable of isolating the subtle, high-frequency oscillatory patterns associated with internal phonological processing.

\begin{table}[ht]
\centering
\caption{Detailed performance comparison on the \textbf{BCIC2020-3} imagined speech recognition task. Values represent the average and standard deviation across five random seeds. \textbf{Bold} indicates the best performance; \underline{underline} indicates the second best.}
\label{tab:results_bcic_2020_3}
\setlength{\tabcolsep}{3pt}
\resizebox{\textwidth}{!}{%
\begin{tabular}{@{}llccccccc@{}}
\toprule
 & & \multicolumn{3}{c}{\textbf{Task-Specific Models}} & \multicolumn{4}{c}{\textbf{Foundation Models}} \\
 \cmidrule(lr){3-5} \cmidrule(lr){6-9}
\textbf{} & \textbf{Sub-metric} & \textbf{EEGNet} & \textbf{EEGConf} & \textbf{FFCL} & \textbf{LaBraM} & \textbf{CBramod} & \textbf{CSBrain} & \textbf{SpecMoE} \\ \midrule
\textbf{\# Params} & --- & --- & --- & --- & 5.8 M & 27.7 M & 31.7 M & 4.3 M \\ \midrule
\multirow{2}{*}{\textbf{Bal. Acc}} & Mean & 0.4413 & 0.4506 & 0.4678 & 0.5060 & 0.5373 & \underline{0.6004} & \textbf{0.6262} \\
 & Std Dev & $\pm$0.0096 & $\pm$0.0133 & $\pm$0.0197 & $\pm$0.0155 & $\pm$0.0108 & $\pm$0.0187 & $\pm$0.0166 \\ \midrule
\multirow{2}{*}{\textbf{Weighted F1}} & Mean & 0.4413 & 0.4488 & 0.4689 & 0.5054 & 0.5383 & \underline{0.6003} & \textbf{0.6264} \\
 & Std Dev & $\pm$0.0102 & $\pm$0.0154 & $\pm$0.0205 & $\pm$0.0205 & $\pm$0.0096 & $\pm$0.0192 & $\pm$0.0158 \\ \bottomrule
\end{tabular}%
}
\end{table}

\subsection{Abnormal Signal Detection Task}
Abnormal detection facilitates the identification of pathological neuronal activity, potentially reducing the clinical workload by providing automated alerts during continuous monitoring. We evaluate our model on the TUAB dataset \cite{obeid2016temple}, a standard benchmark for pathological signal detection. The dataset contains EEG recordings from 23 channels at 256 Hz, annotated as normal or abnormal. Following existing literature, we utilize 16 bipolar montage channels based on the international 10--20 system. Signals were resampled to 200 Hz and segmented into 10-second windows, totaling 409,455 samples. Using the dataset's predefined splits, we further partitioned the training subjects into training and validation sets at an 8:2 ratio. To accelerate the finetuning process and assess model efficiency, we randomly selected a subset of 50,000 samples for training and 50,000 samples for validation.

As shown in Table \ref{tab:results_tuab}, clinical abnormality detection remains a competitive domain for foundation models. SpecMoE maintains a robust stance, reaching a balanced accuracy of \textbf{0.7742} and an AUROC of \textbf{0.8554}. While certain models like CSBrain (\textbf{0.8172}) currently lead the benchmark, it is important to note that our results were achieved using a significantly reduced subset of the available training data. This preliminary result suggests that the spectral-anchored gating mechanism may offer high data efficiency, as the model captured essential pathological signatures despite using only a fraction of the available training data. Extensive tests utilizing the full dataset are planned for future work to further characterize the scaling behavior of SpecMoE in this specific task.

\begin{table}[ht]
\centering
\caption{Detailed performance comparison on the \textbf{TUAB} abnormal signal detection task. Values represent the average and standard deviation across five random seeds. \textbf{Bold} indicates the best performance; \underline{underline} indicates the second best.}
\label{tab:results_tuab}
\setlength{\tabcolsep}{3pt}
\resizebox{\textwidth}{!}{%
\begin{tabular}{@{}llccccccc@{}}
\toprule
 & & \multicolumn{3}{c}{\textbf{Task-Specific Models}} & \multicolumn{4}{c}{\textbf{Foundation Models}} \\
 \cmidrule(lr){3-5} \cmidrule(lr){6-9}
\textbf{} & \textbf{Sub-metric} & \textbf{EEGNet} & \textbf{EEGConf} & \textbf{FFCL} & \textbf{LaBraM} & \textbf{CBramod} & \textbf{CSBrain} & \textbf{SpecMoE} \\ \midrule
\textbf{\# Params} & --- & --- & --- & --- & 5.8 M & 24.4 M & 28.4 M & 4.3 M \\ \midrule
\multirow{2}{*}{\textbf{Bal. Acc}} & Mean & 0.7642 & 0.7758 & 0.7848 & \underline{0.8140} & 0.7891 & \textbf{0.8172} & 0.7742 \\
 & Std Dev & $\pm$0.0036 & $\pm$0.0049 & $\pm$0.0038 & $\pm$0.0019 & $\pm$0.0030 & $\pm$0.0043 & $\pm$0.0070 \\ \midrule
\multirow{2}{*}{\textbf{AUPRC}} & Mean & 0.8299 & 0.8427 & 0.8448 & \underline{0.8965} & 0.8636 & \textbf{0.9005} & 0.8548 \\
 & Std Dev & $\pm$0.0043 & $\pm$0.0054 & $\pm$0.0065 & $\pm$0.0016 & $\pm$0.0063 & $\pm$0.0066 & $\pm$0.0092 \\ \midrule
\multirow{2}{*}{\textbf{AUROC}} & Mean & 0.8412 & 0.8445 & 0.8569 & \textbf{0.9022} & 0.8606 & \underline{0.8957} & 0.8554 \\
 & Std Dev & $\pm$0.0031 & $\pm$0.0038 & $\pm$0.0051 & $\pm$0.0009 & $\pm$0.0057 & $\pm$0.0046 & $\pm$0.0082 \\ \bottomrule
\end{tabular}%
}
\end{table}

\subsection{Seizure Detection Task}
The Siena dataset \cite{PhysioNet-siena-scalp-eeg-1.0.0} is a clinical database comprising video-EEG monitoring from 14 adult patients. EEG signals were recorded at 512 Hz using the international 10--20 system, with seizure events rigorously annotated by clinical experts following the International League Against Epilepsy (ILAE) criteria. We utilized the 29 EEG channels consistently available across the cohort and resampled the signals to 200 Hz. The data was segmented into 10-second windows, totaling 51,307 samples. To ensure a robust evaluation of clinical generalization, we employed a subject-independent split: data from subjects PN16 and PN17 were held out for testing, while the remaining 12 subjects were used for training and validation (8:2 ratio).

As demonstrated in Table \ref{tab:results_siena}, SpecMoE achieves remarkable performance on the Siena benchmark, significantly outperforming all task-specific and foundation models. Notably, SpecMoE reaches a balanced accuracy of \textbf{0.8655} and an AUPRC of \textbf{0.9906}. This represents a substantial lead of \textbf{9.93\%} in balanced accuracy and a \textbf{50.35\%} improvement in AUPRC over the previous state-of-the-art, CSBrain. While such a high AUPRC suggests nearly perfect identification of seizure events within this specific subject-independent split, these results underscore the power of spectral-anchored gating in isolating the distinct rhythmic discharges that characterize ictal activity. The hierarchical U-shaped architecture of SpecHi-Net appears particularly well-suited for capturing these multi-scale temporal dependencies, providing a clear advantage in clinical seizure monitoring.

\begin{table}[ht]
\centering
\caption{Detailed performance comparison on the \textbf{SIENA} seizure detection task. Values represent the average and standard deviation across five random seeds. \textbf{Bold} indicates the best performance; \underline{underline} indicates the second best.}
\label{tab:results_siena}
\setlength{\tabcolsep}{3pt}
\resizebox{\textwidth}{!}{%
\begin{tabular}{@{}llccccccc@{}}
\toprule
 & & \multicolumn{3}{c}{\textbf{Task-Specific Models}} & \multicolumn{4}{c}{\textbf{Foundation Models}} \\
 \cmidrule(lr){3-5} \cmidrule(lr){6-9}
\textbf{} & \textbf{Sub-metric} & \textbf{EEGNet} & \textbf{EEGConf} & \textbf{FFCL} & \textbf{LaBraM} & \textbf{CBramod} & \textbf{CSBrain} & \textbf{SpecMoE} \\ \midrule
\textbf{\# Params} & --- & --- & --- & --- & 5.8 M & 37.7 M & 41.7 M & 4.3 M \\ \midrule
\multirow{2}{*}{\textbf{Bal. Acc}} & Mean & 0.7487 & 0.7556 & 0.6616 & 0.7082 & 0.7317 & \underline{0.7662} & \textbf{0.8655} \\
 & Std Dev & $\pm$0.0521 & $\pm$0.0210 & $\pm$0.0391 & $\pm$0.0329 & $\pm$0.0647 & $\pm$0.0471 & $\pm$0.0038 \\ \midrule
\multirow{2}{*}{\textbf{AUPRC}} & Mean & 0.3753 & 0.2091 & 0.3938 & 0.3122 & 0.4107 & \underline{0.4871} & \textbf{0.9906} \\
 & Std Dev & $\pm$0.0867 & $\pm$0.0786 & $\pm$0.0903 & $\pm$0.0976 & $\pm$0.0720 & $\pm$0.0343 & $\pm$0.0008 \\ \midrule
\multirow{2}{*}{\textbf{AUROC}} & Mean & 0.8687 & 0.8159 & 0.8154 & 0.8814 & 0.9038 & \underline{0.9076} & \textbf{0.9148} \\
 & Std Dev & $\pm$0.0527 & $\pm$0.0261 & $\pm$0.1155 & $\pm$0.0328 & $\pm$0.0218 & $\pm$0.0119 & $\pm$0.0098 \\ \bottomrule
\end{tabular}%
}
\end{table}

\subsection{Vigilance Estimation Task}
SEED-VIG \cite{1741-2552-14-2-026017} is a specialized dataset designed for the continuous estimation of driver vigilance. The data was collected using a virtual driving simulator where 21 subjects performed a driving task while their vigilance levels were continuously monitored. Ground-truth vigilance labels were derived from eye-tracking data using the PERCLOS (Percentage of Closure) indicator. EEG signals were recorded from 17 channels at 200 Hz and segmented into 20,355 8-second windows. We follow a subject-independent split protocol, utilizing subjects 1--13 for training, 14--17 for validation, and 18--21 for testing.

As shown in Table \ref{tab:results_seed_vig}, vigilance estimation serves as a rigorous test for regression-based neural decoding. SpecMoE achieves the highest overall performance in terms of error minimization and variance explanation, reaching a state-of-the-art \textbf{RMSE of 0.1522} and an \textbf{$R^2$ score of 0.2454}. This represents a substantial \textbf{0.1252 reduction in RMSE} compared to the previous best foundation model, CSBrain (\textbf{0.2774}). While LaBraM and CSBrain maintain a narrow lead in Pearson’s correlation coefficient, the significantly lower RMSE of SpecMoE indicates that our spectral-anchored gating mechanism is more effective at minimizing large predictive deviations.

\begin{table}[ht]
\centering
\caption{Detailed performance comparison on the \textbf{SEED-VIG} vigilance estimation (regression) task. Values represent the average and standard deviation across five random seeds. \textbf{Bold} indicates the best performance; \underline{underline} indicates the second best. For RMSE, lower values are better.}
\label{tab:results_seed_vig}
\setlength{\tabcolsep}{3pt}
\resizebox{\textwidth}{!}{%
\begin{tabular}{@{}llccccccc@{}}
\toprule
 & & \multicolumn{3}{c}{\textbf{Task-Specific Models}} & \multicolumn{4}{c}{\textbf{Foundation Models}} \\
 \cmidrule(lr){3-5} \cmidrule(lr){6-9}
\textbf{} & \textbf{Sub-metric} & \textbf{EEGNet} & \textbf{EEGConf} & \textbf{FFCL} & \textbf{LaBraM} & \textbf{CBramod} & \textbf{CSBrain} & \textbf{SpecMoE} \\ \midrule
\textbf{\# Params} & --- & --- & --- & --- & 5.8 M & 21.9 M & 25.9 M & 4.3 M \\ \midrule
\multirow{2}{*}{\textbf{Pearson's corr}} & Mean & 0.5127 & 0.5800 & 0.4923 & \textbf{0.6347} & 0.5502 & \underline{0.6314} & 0.5168 \\
 & Std Dev & $\pm$0.0357 & $\pm$0.0174 & $\pm$0.0313 & $\pm$0.0135 & $\pm$0.0115 & $\pm$0.0356 & $\pm$0.0218 \\ \midrule
\multirow{2}{*}{\textbf{$R^2$ Score}} & Mean & 0.1960 & 0.2065 & 0.1740 & 0.1808 & 0.0737 & \underline{0.2363} & \textbf{0.2454} \\
 & Std Dev & $\pm$0.0427 & $\pm$0.0230 & $\pm$0.0530 & $\pm$0.0958 & $\pm$0.0167 & $\pm$0.0519 & $\pm$0.0289 \\ \midrule
\multirow{2}{*}{\textbf{RMSE ($\downarrow$)}} & Mean & 0.2847 & 0.2829 & 0.2885 & 0.2871 & 0.3057 & \underline{0.2774} & \textbf{0.1522} \\
 & Std Dev & $\pm$0.0076 & $\pm$0.0041 & $\pm$0.0093 & $\pm$0.0166 & $\pm$0.0027 & $\pm$0.0094 & $\pm$0.0029 \\ \bottomrule
\end{tabular}%
}
\end{table}

\section{Additional results}
\subsection{Detailed Ablation Studies}
\begin{figure}
    \centering
    \includegraphics[width=1\linewidth]{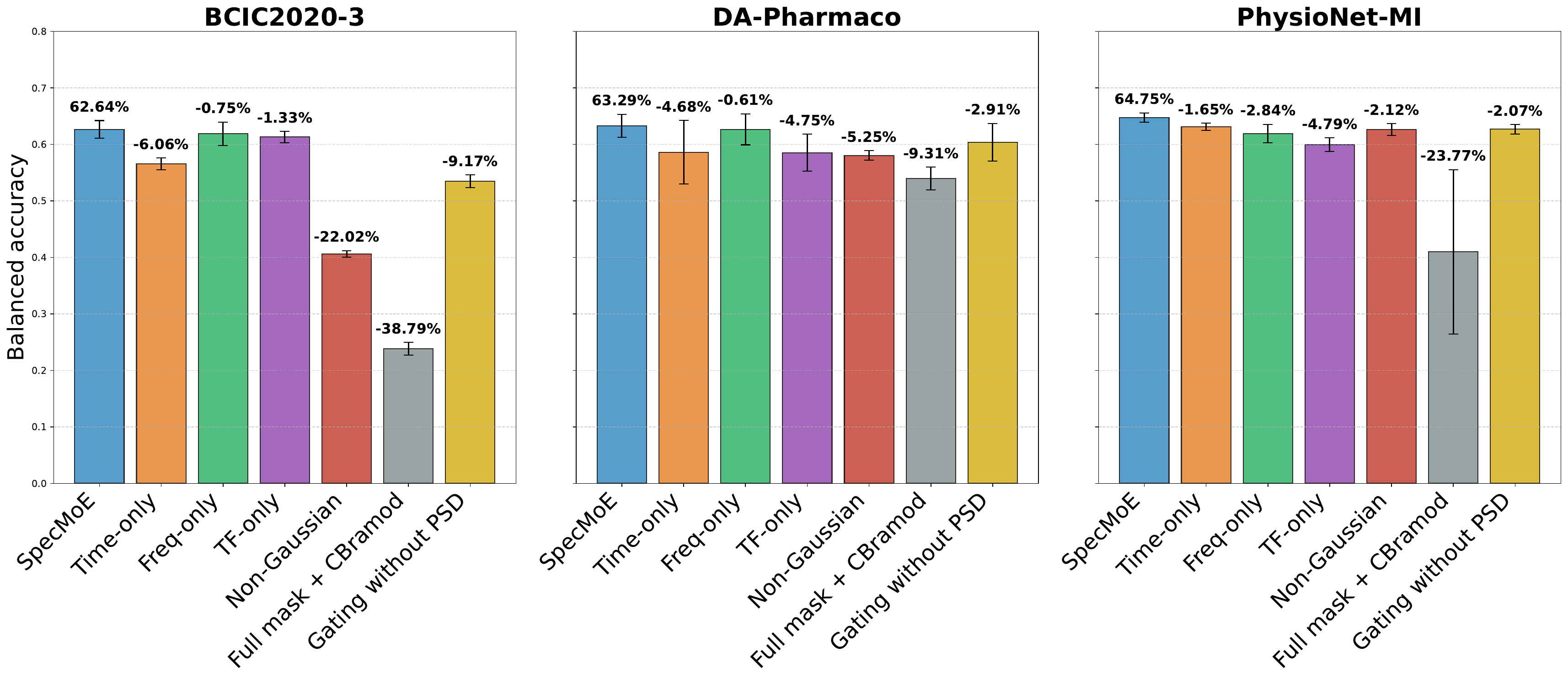}
    \caption{SpecMoE ablation results - balanced accuracy. We show the absolute value of the balanced accuracy for the SpecMoE model and relative differences for six ablation experiments. TF stands for time-frequency.}
    \label{fig:Ablation_Study_Balanced_Acc}
\end{figure}

\begin{figure}
    \centering
    \includegraphics[width=1\linewidth]{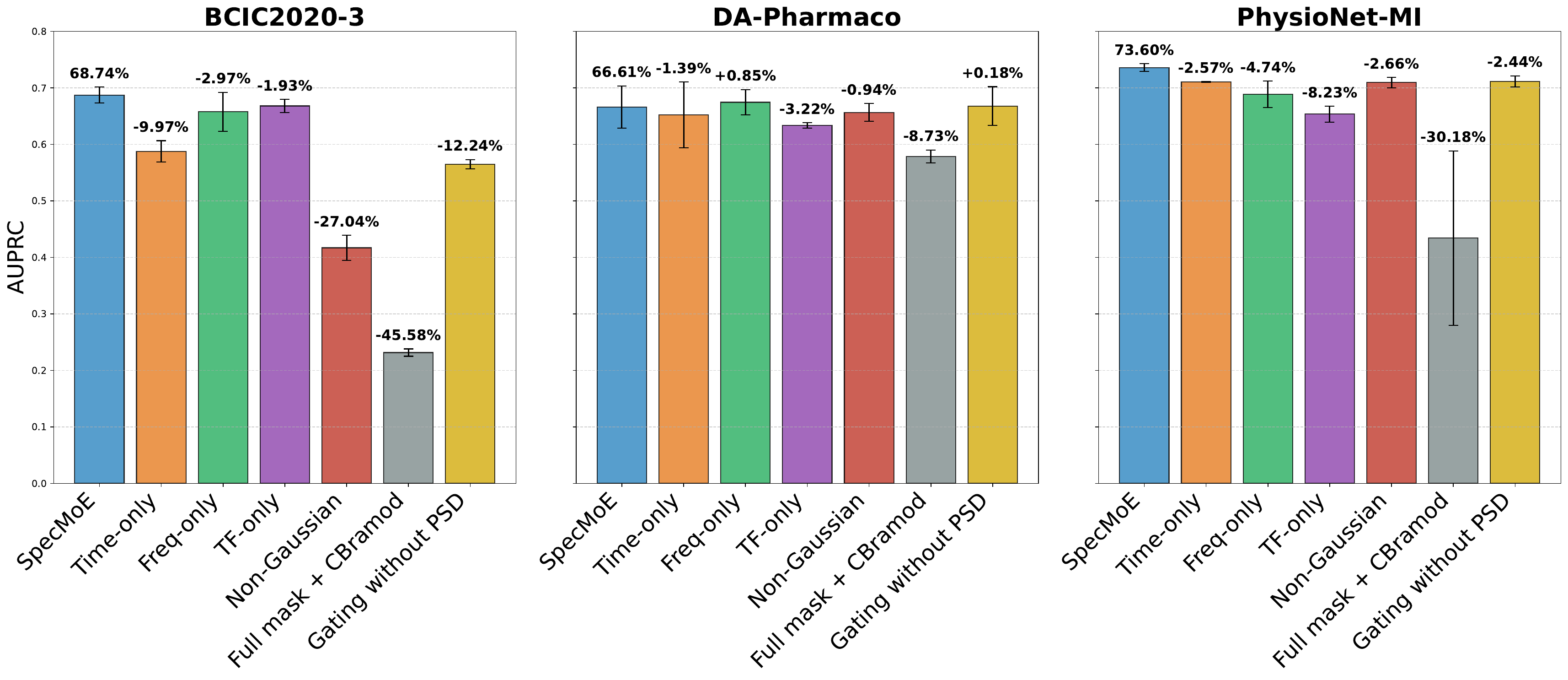}
    \caption{SpecMoE ablation results - AUPRC. We show the absolute value of AUPRC for the SpecMoE model and relative differences for six ablation experiments. TF stands for time-frequency.}
    \label{fig:Ablation_Study_AUPRC}
\end{figure}

\begin{figure}
    \centering
    \includegraphics[width=1\linewidth]{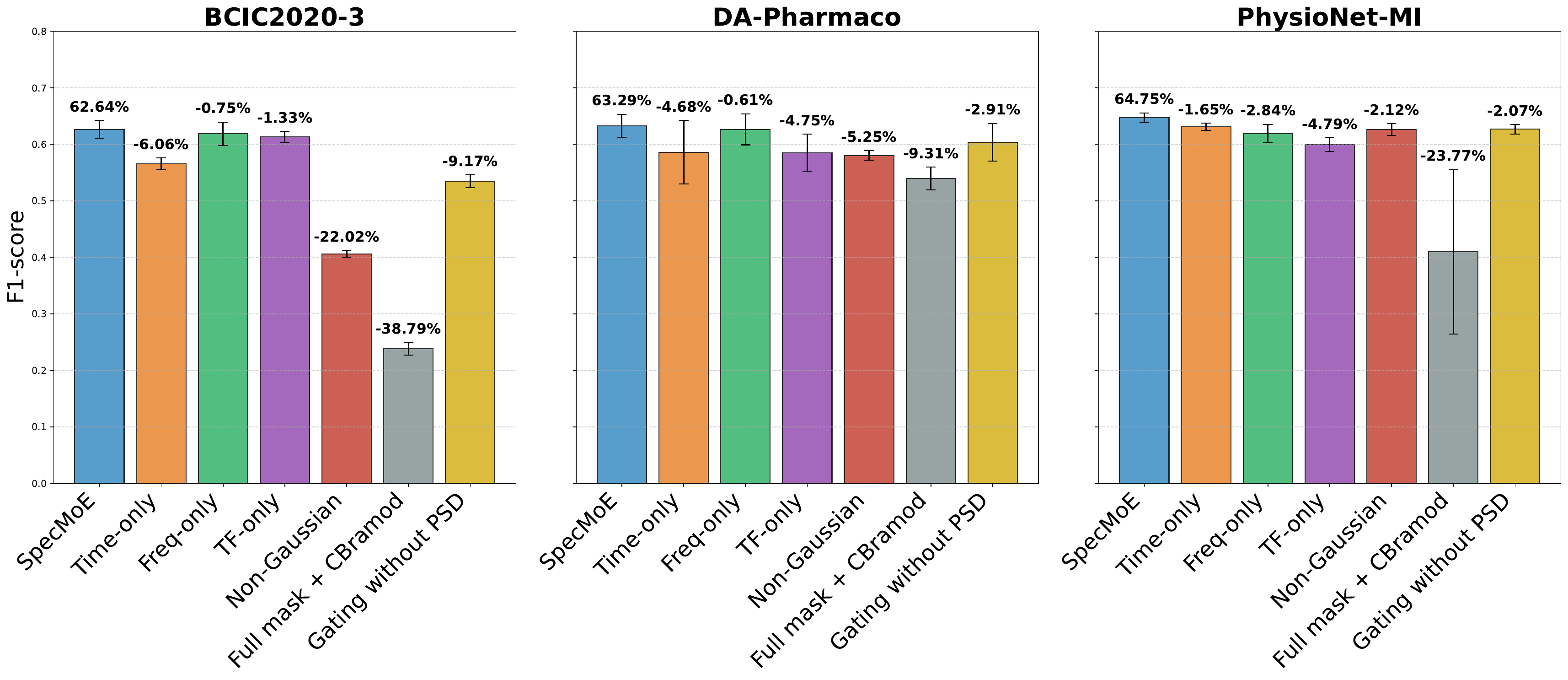}
    \caption{SpecMoE ablation results - F1-score. We show the absolute value of F1-score for the SpecMoE model and relative differences for six ablation experiments. TF stands for time-frequency.}
    \label{fig:Ablation_Study_F1_Score}
\end{figure}

Figures \ref{fig:Ablation_Study_Balanced_Acc}, \ref{fig:Ablation_Study_AUPRC}, and \ref{fig:Ablation_Study_F1_Score} illustrate the relative impact of our core design choices across the BCIC2020-3, DA-Pharmaco, and PhysioNet-MI datasets. By systematically removing or replacing key components, we demonstrate that the high performance of SpecMoE is derived from the synergy between our novel masking strategy and the spectral-guided mixture-of-experts.

\paragraph{Impact of Gaussian-Smoothed Masking.}
The most critical component of our framework is the Gaussian-smoothed masking scheme. When replaced with standard \textbf{Non-Gaussian (rectangular) masks}, we observe a substantial drop in performance across all tasks. For instance, on BCIC2020-3, the balanced accuracy plummets from 0.6262 to 0.4063 (a relative decrease of approximately 22\%). This experiment confirms our hypothesis that "soft" Gaussian boundaries force the model to learn the complex neural oscillations from a large spectral context.

\paragraph{Efficacy of Joint Time-Frequency (TF) Masking.}
We evaluated the necessity of joint masking by comparing it against Time-only and Frequency-only variants. On the DA-Pharmaco dataset, utilizing only time masks resulted in an F1-score drop from 0.6329 to 0.5861. The consistent superiority of the full TF-masking approach across all metrics suggests that simultaneous occlusion in both domains is essential for capturing the non-stationary rhythmic signatures of EEG signals, particularly in pharmacological and clinical benchmarks.

\paragraph{Architecture and Gating Strategy.}
The ablation labeled "Full mask + CBramod" replaces our hierarchical SpecHi-Net and MoE structure with the CBramod backbone while keeping the Gaussian masking. The results show a massive decline (e.g., a drop to 0.2448 balanced accuracy on BCIC2020-3), indicating that non-hierarchical architectures lack the capacity to effectively solve the challenging reconstruction tasks posed by our aggressive masking strategy.

Furthermore, the "Gating without PSD" experiment removes the Power Spectral Density prior from the routing network. This leads to a consistent performance degradation—most notably on BCIC2020-3, where the AUPRC drops by over 12\% (from 0.6874 to 0.5650). This validates that the PSD serves as a vital physical anchor, enabling the gating network to route signals to specialized experts based on their underlying rhythmic content rather than mere temporal patterns.

\subsection{Masking Effects Visualization on STFT Spectrograms}
\begin{figure}
    \centering
    \includegraphics[width=1\linewidth]{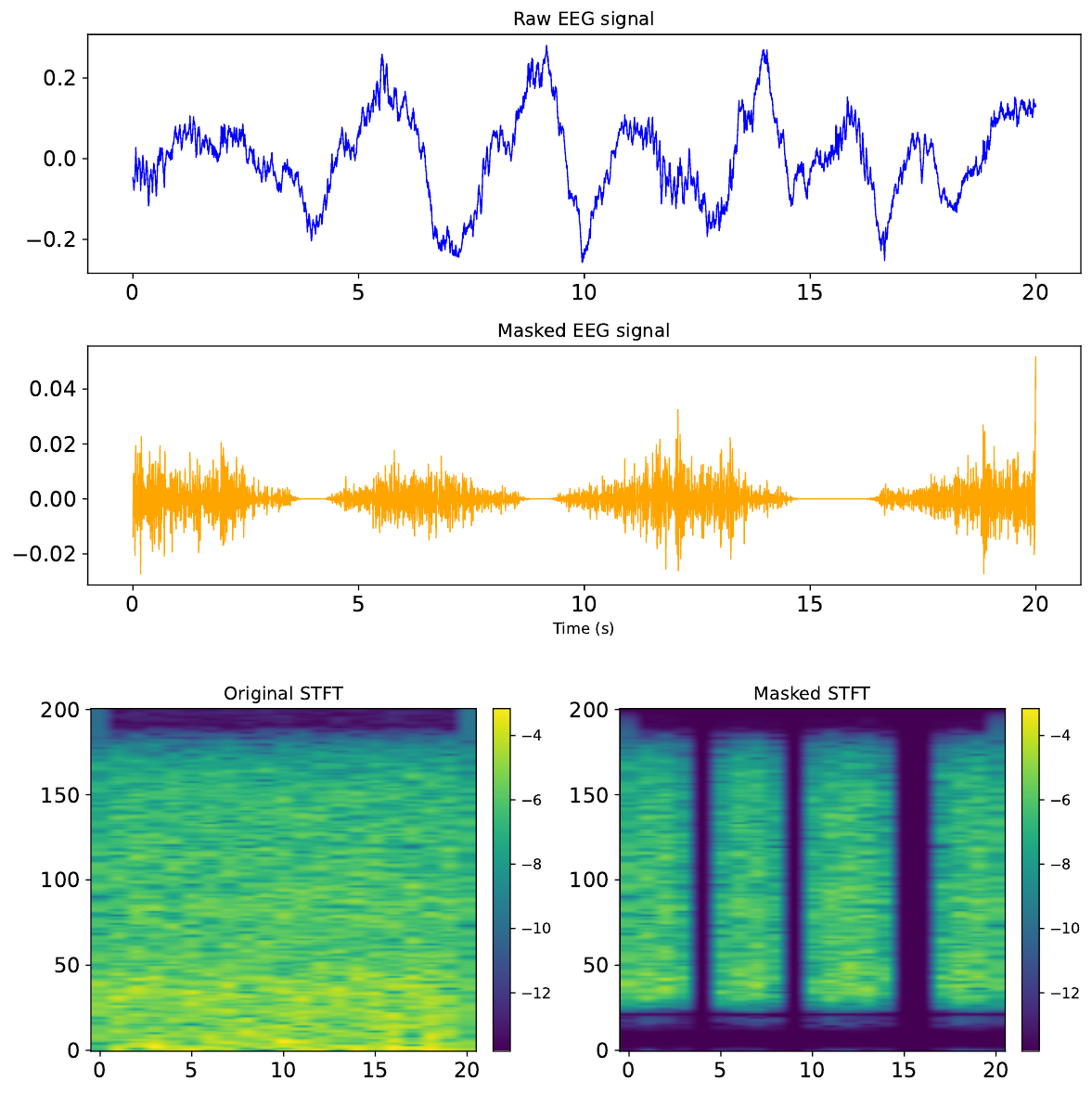}
    \caption{Overview of the proposed joint Gaussian masking strategy. The panels illustrate the simultaneous application of temporal, spectral, and joint time-frequency occlusions on raw EEG signals (top panel: raw, middle panel: masked) and their corresponding STFT spectrograms (bottom panel).}
    \label{fig:TUEG_masking_V1}
\end{figure}

\begin{figure}
    \centering
    \includegraphics[width=1\linewidth]{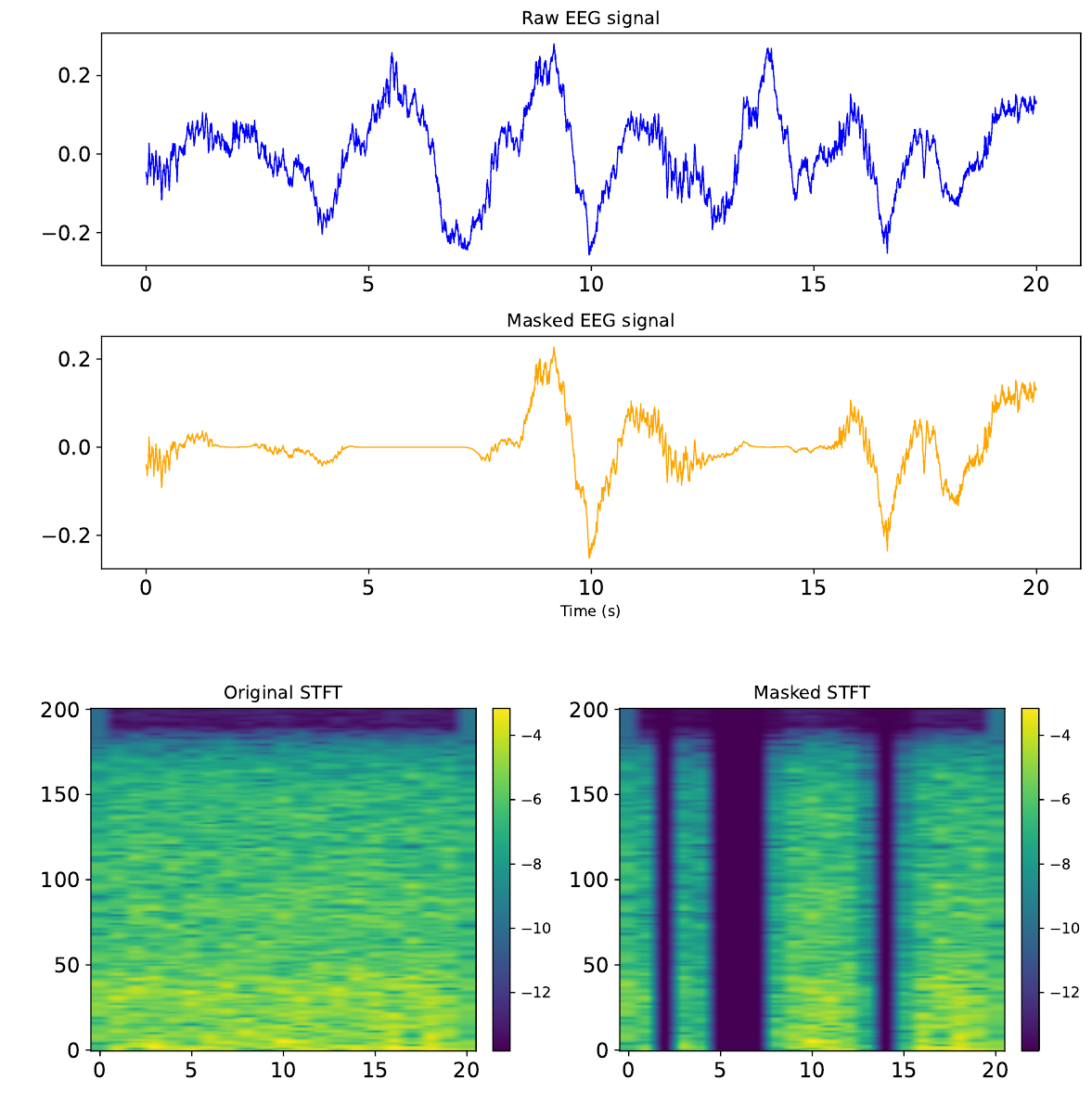}
    \caption{Visualization of temporal domain masking.
    The panels illustrate the application of time-based occlusions on raw EEG signals (top panel: raw, middle panel: masked) and their corresponding STFT spectrograms (bottom panel).}
    \label{fig:TUEG_masking_V2}
\end{figure}

\begin{figure}
    \centering
    \includegraphics[width=1\linewidth]{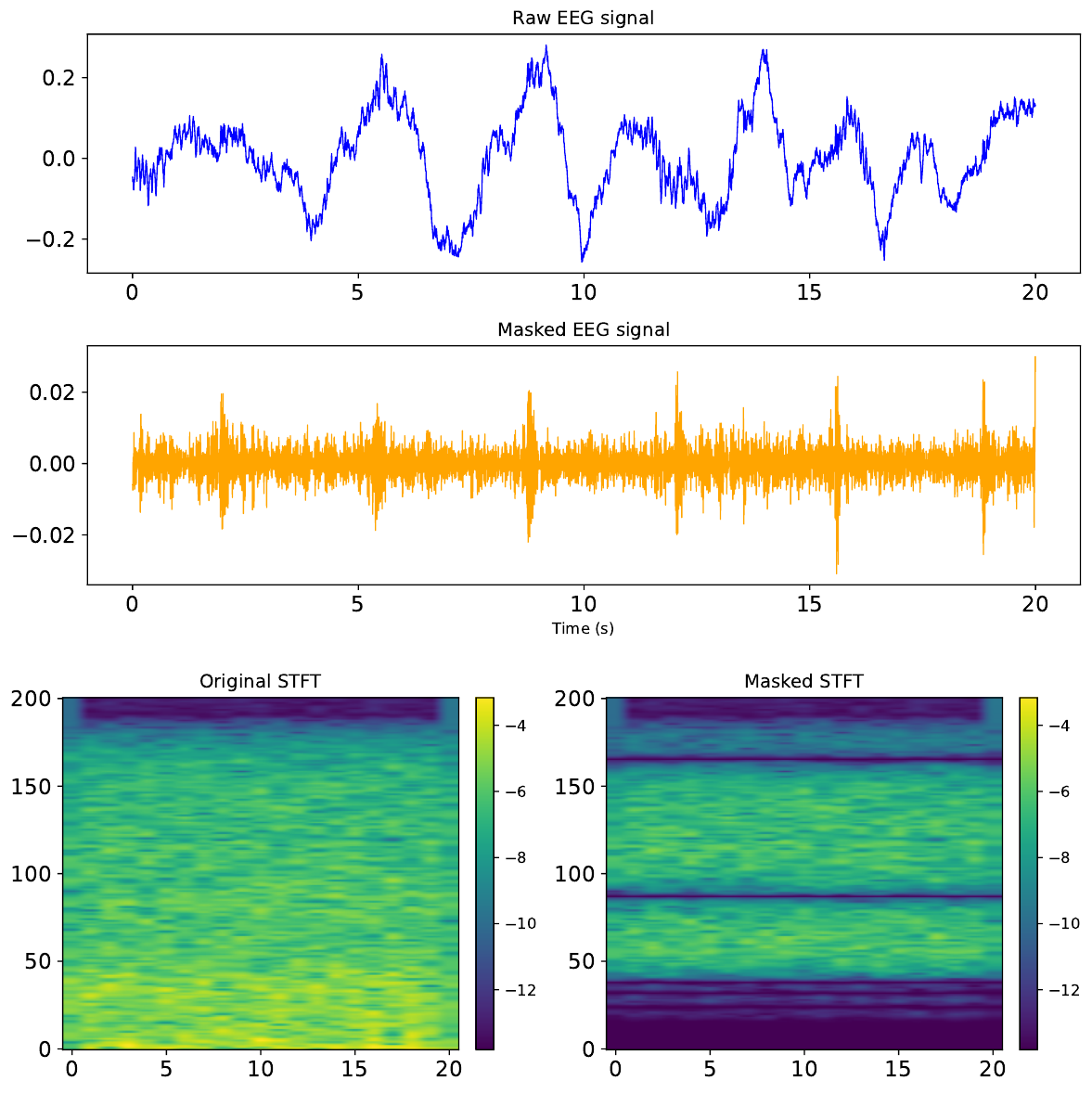}
    \caption{Visualization of frequency domain masking. The panels illustrate the application of spectral-based occlusions on raw EEG signals (top panel: raw, middle panel: masked) and their corresponding STFT spectrograms (bottom panel).}
    \label{fig:TUEG_masking_V3}
\end{figure}

\begin{figure}
    \centering
    \includegraphics[width=1\linewidth]{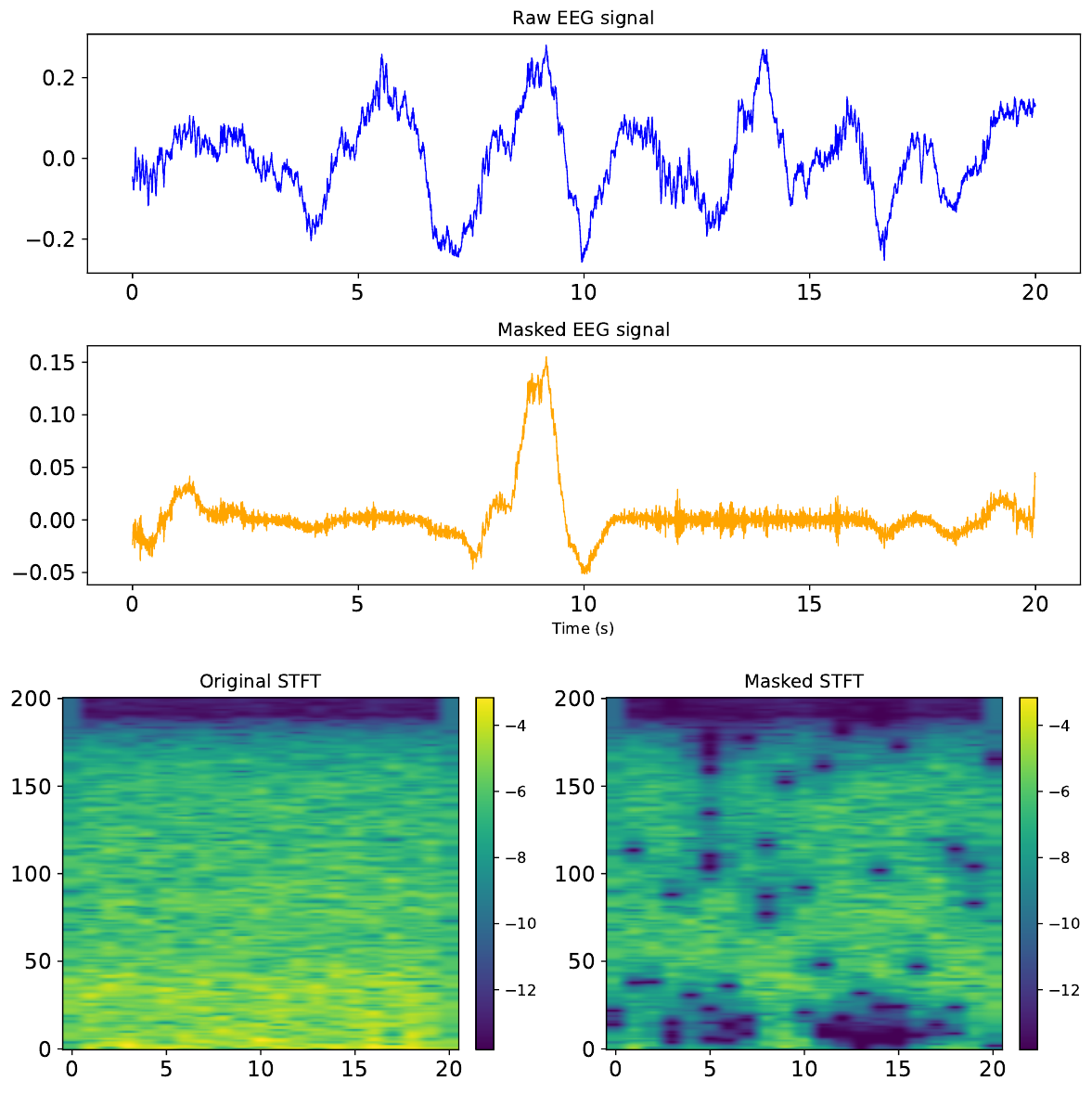}
    \caption{Visualization of time-frequency domain masking.
    The panels illustrate the application of temporal-spectral blobs occlusions on raw EEG signals (top panel: raw, middle panel: masked) and their corresponding STFT spectrograms (bottom panel).}
    \label{fig:TUEG_masking_V4}
\end{figure}

\begin{figure}
    \centering
    \includegraphics[width=1\linewidth]{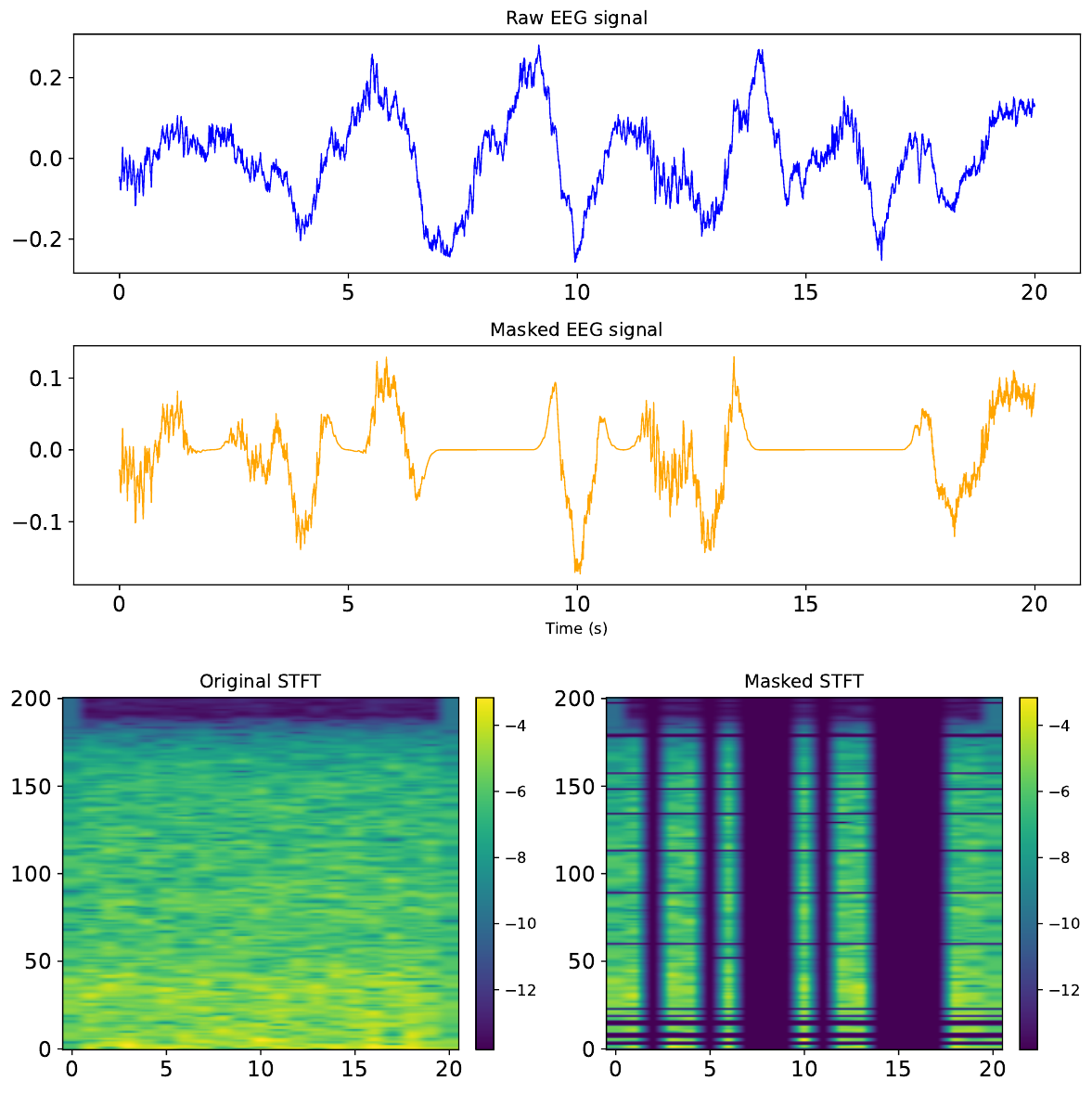}
    \caption{Time, frequency and time-frequency masking with rectangular masks. 
    The panels illustrate the simultaneous application of temporal, spectral, and joint time-frequency rectangular occlusions on raw EEG signals (top panel: raw, middle panel: masked) and their corresponding STFT spectrograms (bottom panel).}
    \label{fig:TUEG_masking_V5}
\end{figure}

Figures \ref{fig:TUEG_masking_V1}, \ref{fig:TUEG_masking_V2}, \ref{fig:TUEG_masking_V3}, \ref{fig:TUEG_masking_V4} and  \ref{fig:TUEG_masking_V5} illustrate how the raw EEG signals and their corresponding STFT spectrograms are modified under different masking configurations. These visualizations contrast our proposed Gaussian-smoothed masks (Figures \ref{fig:TUEG_masking_V1}--\ref{fig:TUEG_masking_V4}) against standard rectangular masks (Figure \ref{fig:TUEG_masking_V5}), highlighting the different levels of information occlusion across the time and frequency domains.

\subsection{Pretraining Importance Analysis}
\begin{figure}
    \centering
    \includegraphics[width=1\linewidth]{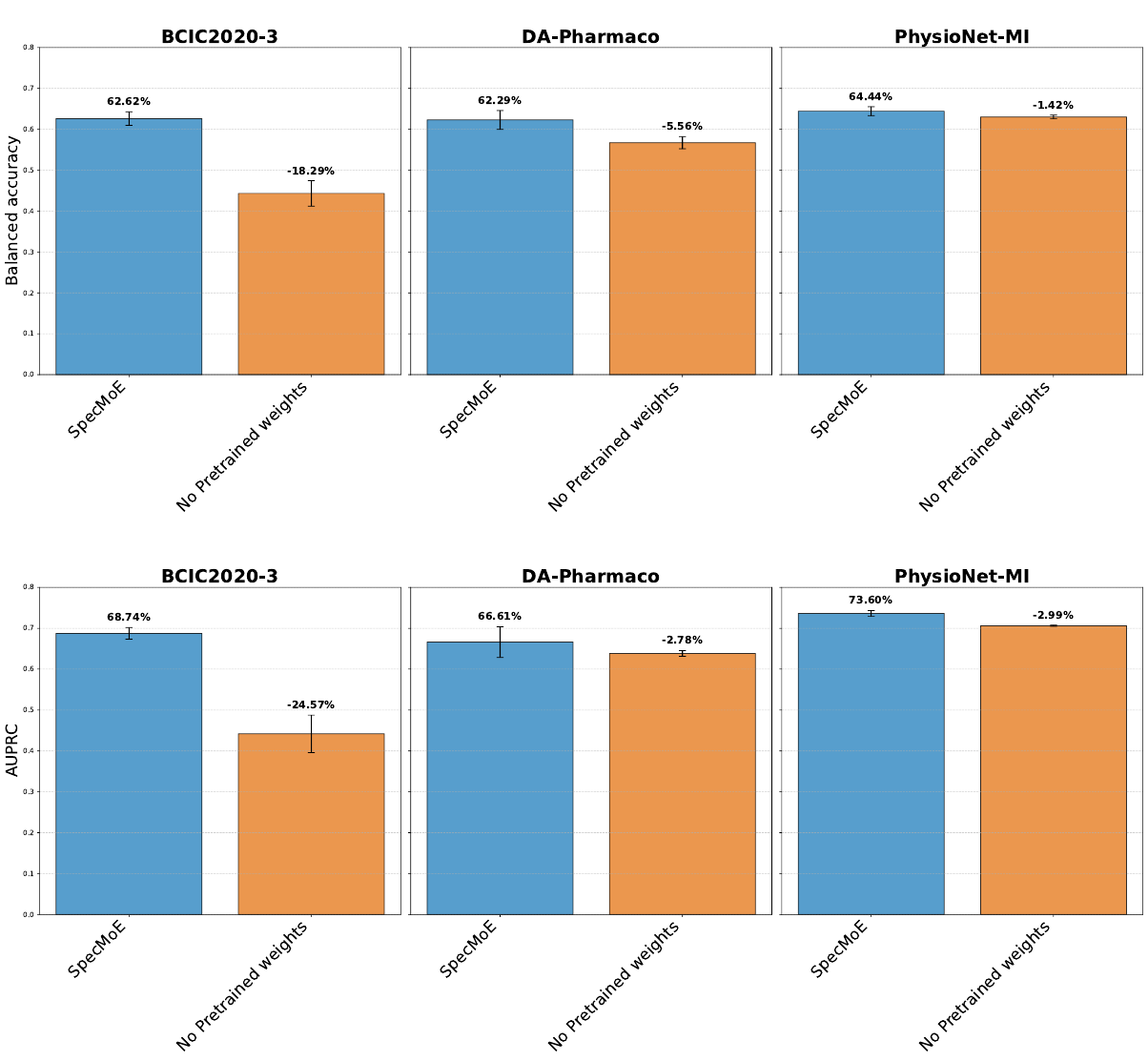}
    \caption{SpecMoE pretraining ablations on six datasets.}
    \label{fig:Pretraining_Ablation}
\end{figure}

Figure \ref{fig:Pretraining_Ablation} illustrates the critical role of self-supervised pretraining by comparing the performance of SpecMoE initialized with pretrained weights against a version trained from scratch (random initialization). Across all three representative datasets—BCIC2020-3, DA-Pharmaco, and PhysioNet-MI—the pretrained model consistently achieves superior results in both balanced accuracy and AUPRC.

The most substantial impact is observed in the \textbf{BCIC2020-3} imagined speech task, where balanced accuracy falls from 0.6262 to 0.4433 without pretraining—an absolute decrease of 18.29 points. Similarly, the AUPRC for this task experienced a sharp decline of 0.2457. This indicates that decoding high-complexity tasks like imagined speech relies heavily on the generalized neural representations learned during large-scale pretraining.

In the \textbf{DA-Pharmaco} and \textbf{PhysioNet-MI} datasets, we observe absolute accuracy gains of 5.56 and 1.42 points, respectively. The improvement in DA-Pharmaco is particularly noteworthy as it validates the \textbf{cross-species} utility of our foundation model; despite being pretrained on human EEG, the model learns universal spectral features that significantly accelerate and improve the decoding of rodent LFP signals. Overall, these results confirm that our Gaussian-smoothed masking task successfully forces the model to learn a robust, generalized representation of brain activity that serves as a high-quality initialization for diverse downstream applications.

\subsection{Experts Contribution Analysis}
\begin{figure}
    \centering
    \includegraphics[width=1\linewidth]{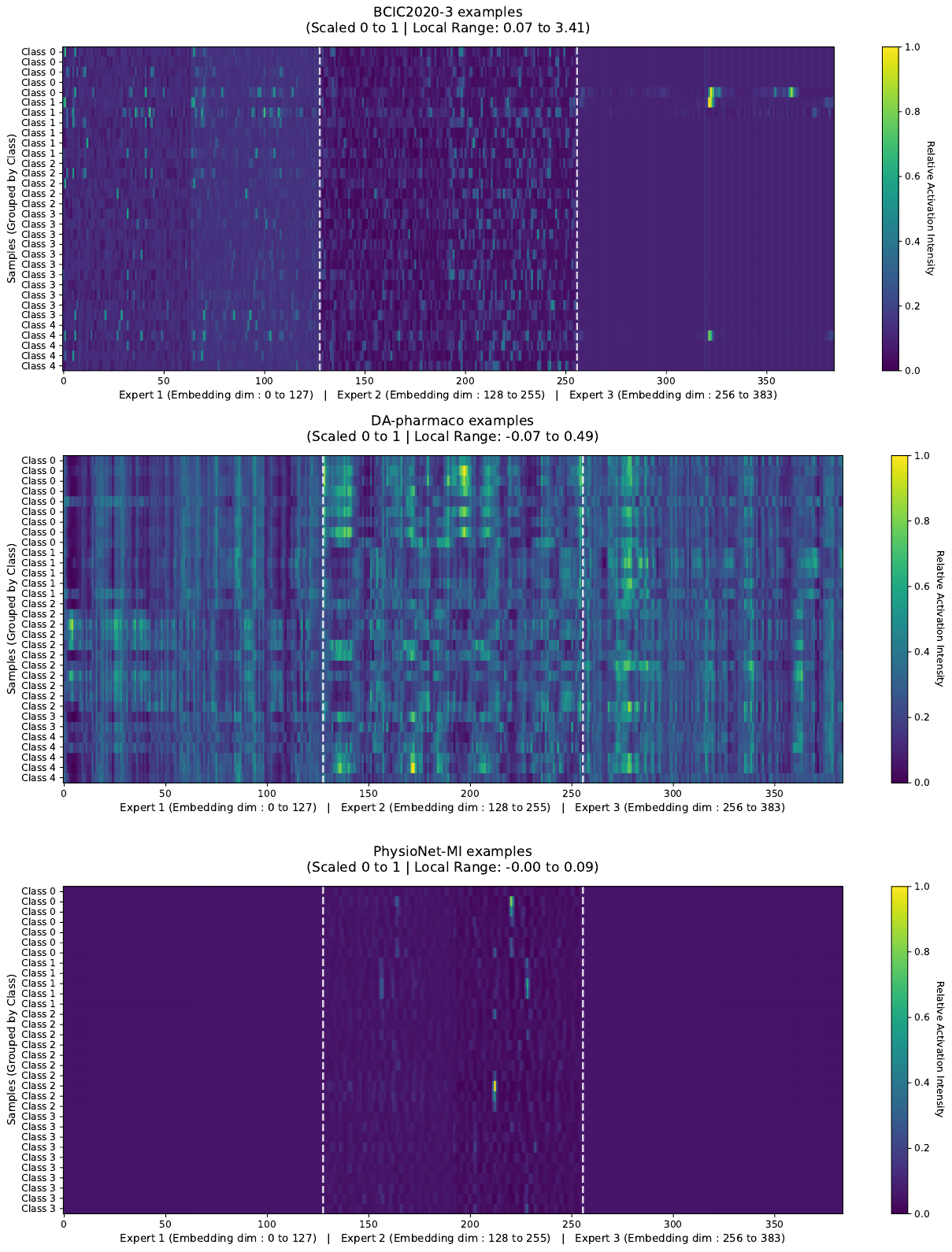}
    \caption{Visualization of expert contribution across different EEG paradigms: BCIC2020-3 (top), DA-Pharmaco (middle), and PhysioNet-MI (bottom). Each panel displays the routing activations for Experts 1, 2, and 3 (from left to right), with samples grouped by class labels. The color intensity represents the gating network's routing weights, ranging from low (darker tones) to high (brighter tones), reflecting the relative importance of each expert for a given sample.}
    \label{fig:Experts_Contribution}
\end{figure}

Figure \ref{fig:Experts_Contribution} visualizes the relative contribution of each expert across three datasets, revealing distinct routing patterns tailored to each task's complexity.

On the BCIC2020-3 dataset, we observe that the gating network primarily solicits Experts 1 and 2. This shared contribution suggests that the decoding of imagined speech, a high-complexity task, requires a collaborative representation where different experts likely specialize in complementary spectral features of internal phonological processing.

In another task, the DA-Pharmaco dataset engages all three experts across the different classes and brain regions. This broad utilization reflects the high variance of the pharmacological signatures in murine LFP data, where different experts are required to isolate the unique oscillatory shifts induced by diverse compounds.

Finally, for PhysioNet-MI, the routing logic is highly sparse, with Expert 2 being almost exclusively solicited. This indicates that for standard motor imagery tasks, the gating network identifies a consistent, dominant spectral profile across the motor cortex, which can be effectively modeled by a single specialized expert without the need for additional parameters from the others.

\subsection{Embeddings Visualization}
We qualitatively assess the model's representation space using t-SNE projections on the DA-Pharmaco and MACO test sets.

\begin{figure}
    \centering
    \includegraphics[width=1\linewidth]{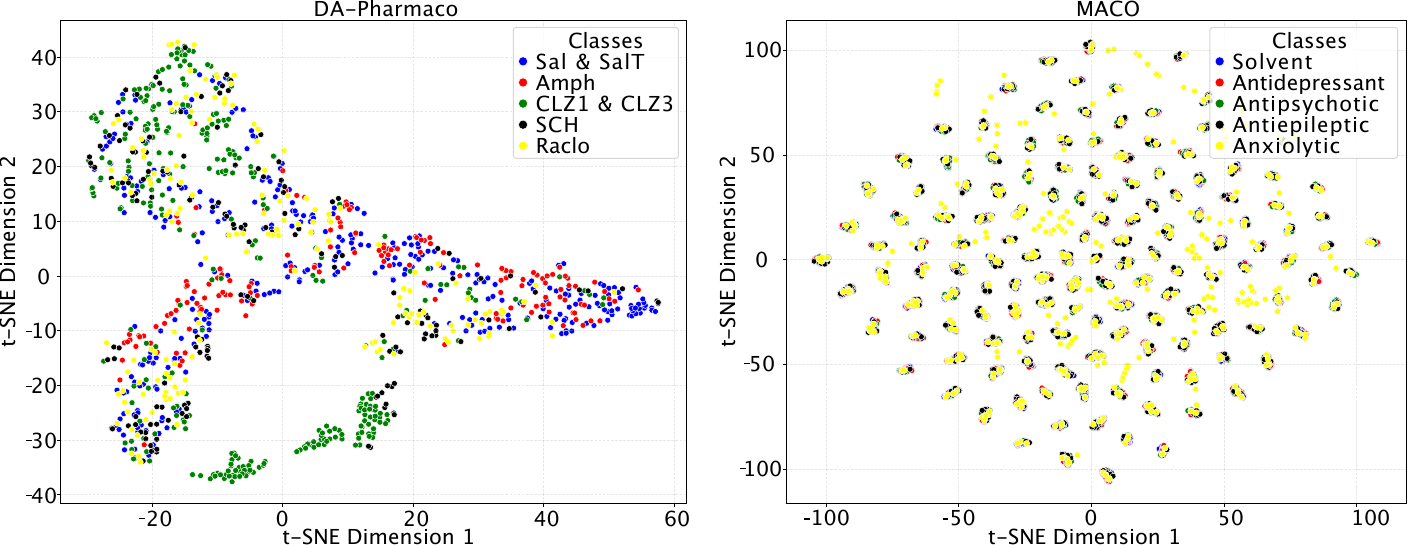}
    \caption{T-SNE DA-Pharmaco (left) and MACO (right) projections obtained by an untrained SpecMoE model containing random weights.
    }
    \label{fig:Pharmaco_EEG_AND_MACO_raw_EEG_projection}
\end{figure}

Figure \ref{fig:Pharmaco_EEG_AND_MACO_raw_EEG_projection} displays the projections obtained using random weights from an untrained SpecMoE model. More specifically, we initialized the model with random weights and projected the output results without any training. For both datasets, the classes are entirely overlapped, confirming that raw signals lack intrinsic linear separability and highlighting the necessity for training.

\begin{figure}
    \centering
    \includegraphics[width=1\linewidth]{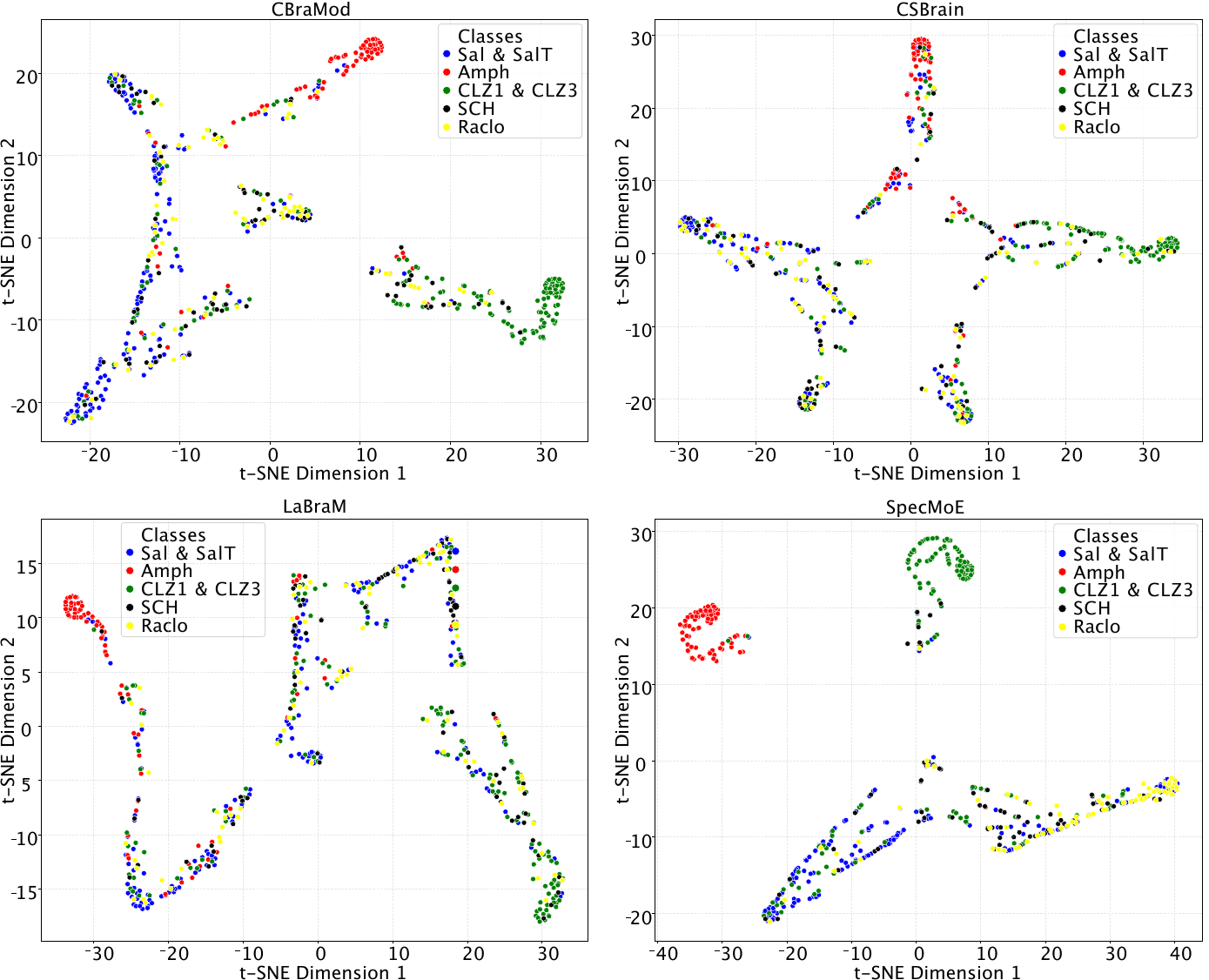}
    \caption{T-SNE DA-Pharmaco test data projection using four foundation models. Top left: CBraMod, top right: CSBrain, bottom left: LaBraM, bottom right: SpecMoE.}
    \label{fig:Pharmaco_EEG_Embeddings}
\end{figure}

\begin{figure}
    \centering
    \includegraphics[width=1\linewidth]{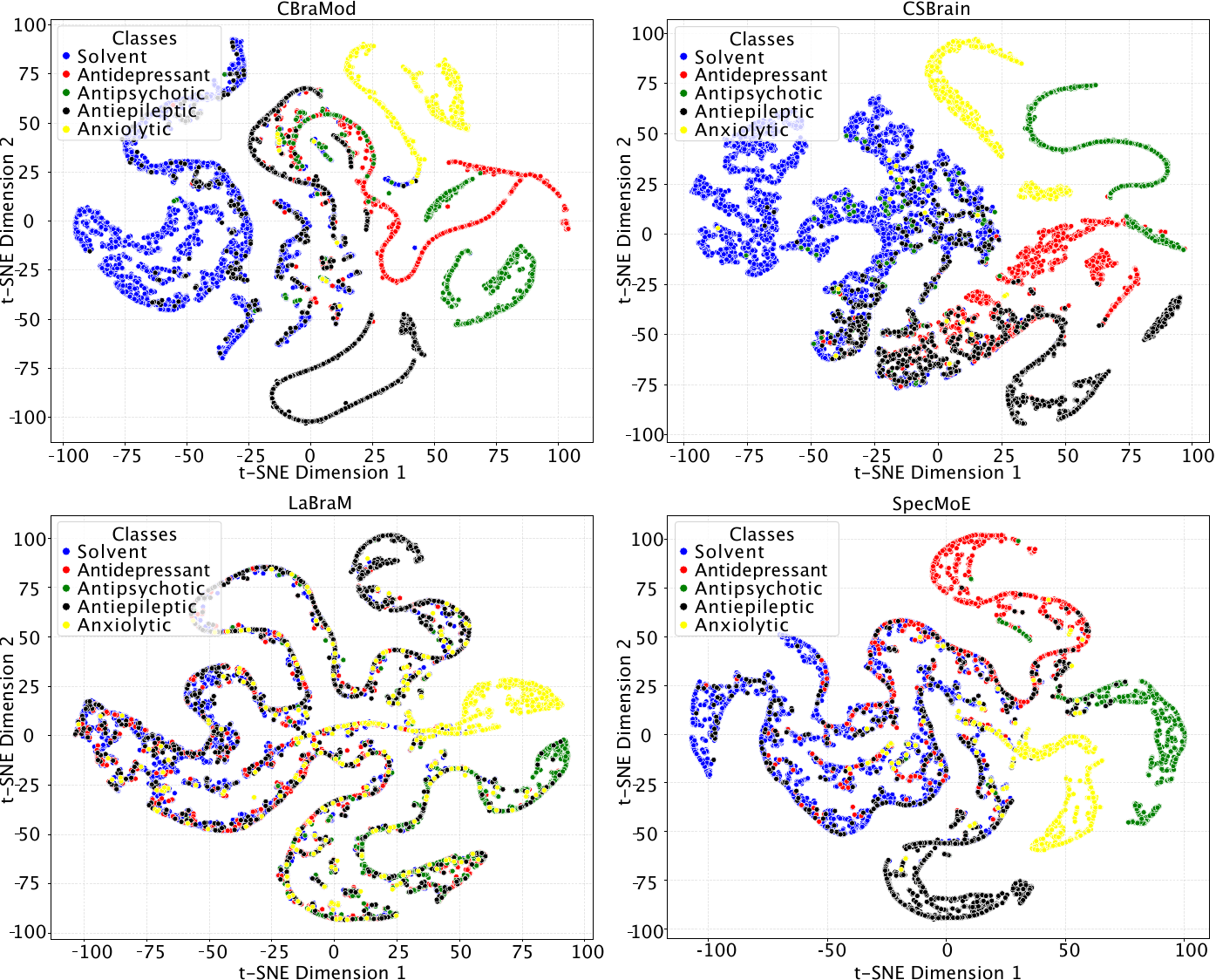}
    \caption{T-SNE MACO test data projection using four foundation models. Top left: CBraMod, top right: CSBrain, bottom left: LaBraM, bottom right: SpecMoE.}
    \label{fig:MACO_Embeddings}
\end{figure}

Figures \ref{fig:Pharmaco_EEG_Embeddings} and \ref{fig:MACO_Embeddings} compare SpecMoE against baseline foundation models. On DA-Pharmaco, SpecMoE produces more compact clusters, particularly for the Saline and Amphetamine groups. On MACO, the distinction is even more pronounced: SpecMoE is the only model capable of clearly isolating the Antiepileptic class. These results visually confirm that SpecMoE effectively structures the latent space, mapping diverse pharmacological effects into distinct neural signatures.

\end{document}